\documentclass[a4paper,3p,review]{elsarticle}

\usepackage{hyperref}
\usepackage{placeins}

\journal{Journal of \LaTeX\ Templates}

\biboptions{authoryear}

\begin{document}

\begin{frontmatter}

\title{Intelligent Spatial Interpolation-based Frost Prediction Methodology using Artificial Neural Networks with Limited Local Data.}

\author[1,2]{Ian Zhou}
\ead{ian.zhou@student.uts.edu.au}
\author[1,2]{Justin Lipman}
\ead{justin.lipman@uts.edu.au}
\author[1]{Mehran Abolhasan}
\ead{mehran.abolhasan@uts.edu.au}
\author[1,2]{Negin Shariati}
\ead{negin.shariati@uts.edu.au}
\address[1]{University of Technology Sydney, Australia}
\address[2]{Food Agility CRC Ltd, 175 Pitt St., Sydney, NSW, 2000, Australia}

\begin{abstract}
The weather phenomenon of frost poses great threats to agriculture. As recent frost prediction methods are based on on-site historical data and sensors, extra development and deployment time are required for data collection in any new site. The aim of this article is to eliminate the dependency on on-site historical data and sensors for frost prediction methods. In this article, a frost prediction method based on spatial interpolation is proposed. The models use climate data from existing weather stations, digital elevation models surveys, and normalized difference vegetation index data to estimate a target site's next hour minimum temperature. The proposed method utilizes ensemble learning to increase the model accuracy. Climate datasets are obtained from 75 weather stations across New South Wales and Australian Capital Territory areas of Australia. The results show that the proposed method reached a detection rate up to 92.55\%.
\end{abstract}


\end{frontmatter}

%\linenumbers

\section{Introduction}
\label{spat:intro}
% Frost Background, What I want to achieve here
Frost is a devastating and extreme weather phenomenon that could induce critical losses to agriculture-related industries \citep{doi:10.1111/gcb.14479, crimp2019possible}. As frost events are capable of eliminating the crops from the upstream of supply chains, it could affect jobs and businesses along the entire supply chain \citep{snyder2005frosteco}. Fortunately, frost events could be predicted and detected. Recent frost prediction methods are based on machine learning models \citep{8993828}. These models require historical data from a specific site to predict future frost events. However, this requirement of on-site historical data poses a restriction for model construction in new sites without historical data \citep{8993828}. Long periods of time are needed for data collection at these new sites to construct and deploy the machine learning models. After the models are constructed, local sensor motes are needed to feed live data to the prediction models. This is another restriction of recent methods. To overcome these restrictions, this article proposes a frost prediction method based on spatial interpolation techniques, aiming to predict frost for a site without any on-site historical data or sensors.

% Why I chose this model structure
% Spatial Interpolation
Spatial interpolation includes methods that generate or predict spatially continuous data from a few regional sample points \citep{LI2014173}. In \citep{JEFFREY2001309}, spatial interpolation is used to construct daily environmental data of Australia, including daily minimum temperature, using rainfall and climate data from 4600 different locations. This has demonstrated the potential of spatial interpolation in the construction of different environmental parameters in a $0.05^{\circ}$ grid. \citet{SINGHAL2022105317} revealed the application of spatial interpolation in the Northwest Himalayas to estimate temperature and rainfall in 5--10 km resolutions. Another work \citep{APPELHANS201591} compared different spatial interpolation methods for monthly air temperatures at Mt. Kilimanjaro, Tanzania. They found that model averaging neural networks and Artificial Neural Networks (ANNs) are ranked fourth and fifth in accuracy, respectively. ANNs are also the most accurate model type within recent frost prediction techniques. Therefore, in this article, the proposed method is based on multiple ANN models. The proposed method is also compared to the baseline models that represent previous methods relying on local data sources. These baseline models are also ANN models.

% Ensemble Learning
The proposed method is an ensemble learning method that utilizes the existing historical data from 75 weather stations across New South Wales (NSW) and Australian Capital Territory (ACT) in Australia. Several weak predictors or submodels are trained. Each weak predictor corresponds to one of the weather stations. Each of the weak predictors is created using the climate features from one specific station to predict the next hour minimum temperatures of other weather stations. Location features from both stations are also included. The location features include geographical location, elevation from the Digital Elevation Model (DEM), and Normalized Difference Vegetation Index (NDVI), indicating the amount of green vegetation \citep{CARLSON1997241}. The correlation between NDVI and temperature is discovered in different works. Strong positive correlations are revealed between NDVI and minimum temperature \citep{doi:10.1080/01431160010007033, doi:10.1080/01431160210154812}. Therefore, NDVI is used as a hint for the minimum temperature. These location features are also used to aggregate the results of the weak predictors.

\subsection{Related Work}
% Frost Prediction Works and their flaws
% Link back to what I want to achieve with current flaws and extend.
% Contribution
% Description of sections
Recent frost prediction methods can be divided into ``classification methods'' and ``regression methods'' \citep{8993828}. The classification methods predict the probability of frost occurrence in the future. The regression methods predict the future minimum temperature. An example of classification methods is GENIE \citep{CHEVALIER201284}. GENIE classifies predicted future meteorological conditions into five levels of frosts and freezes with expert agrometeorologists' knowledge. This system monitors potential frost events on blueberries and peaches. However, as different crops have different resistance to frost \citep{di2020drought}, this article focuses on regression methods of frost prediction. Along with the regression methods, different triggering temperatures for frost can be applied as a general solution for different types of crops and plants.

The majority of recent frost prediction regression methods \citep{doi:10.1046/j.1467-8470.2003.00235.x, iacono2018performance, 8447058, GHIELMI2006101, 6391830, fuentes2018application} predict the minimum temperature of the next 12--24 hours. The prediction models include, linear regression \citep{doi:10.1046/j.1467-8470.2003.00235.x, iacono2018performance}, random forest \citep{8447058}, and ANNs \citep{GHIELMI2006101, 6391830, fuentes2018application}. The ANNs have the highest accuracy among the other model types \citep{8993828}.

All these methods \citep{doi:10.1046/j.1467-8470.2003.00235.x, iacono2018performance, 8447058, GHIELMI2006101, 6391830, fuentes2018application} depend on local/on-site historical climate data for model training, validation, testing and future operations. This first generates extra development and deployment time to collect local climate datasets. Then, a live sensor system is required to transmit data to the prediction models. The proposed method in this article aims to remove the model dependency on on-site data and sensors. Models can be built from existing historical data from other weather stations, previously surveyed DEM, and NDVI data from satellites. When the proposed system is operating, the required model input sources are also climate data from other weather stations, previously surveyed DEM, and NDVI data from satellites. The proposed method does not require on-site data or sensors. The major contributions of this article are:
\begin{enumerate}
    \item Proposing a spatial interpolation-based frost prediction method.
    \item Eliminating the on-site data/sensor requirement for frost regression prediction methods.
    \item Exploring the performance of an off-site frost prediction method.
\end{enumerate}

The rest of the paper is organized as follows. Section \ref{spat:meth} presents the methodology. This includes the descriptions of data sources, data preprocessing methods, model types, and experiment settings. The experiments compare the performance of the proposed method with the baseline method derived from on-site climate datasets. Then, Section \ref{spat:result} shows the experiment results along with the discussions on limitations. Finally, the paper is concluded in Section \ref{spat:conclude}.

\section{Methodology}
\label{spat:meth}
% Data Source description
% Preprocessing
% Ensemble Structure
% Experiments
In this section, the data processing and experiment procedures are presented. To start all experiments, multiple datasets are obtained from different data sources. These datasets and their sources are described. Then, the data preprocessing steps are demonstrated, followed by the model structure of the prediction models. Finally, this section explains the experiment settings.

\subsection{Data Sources}
% Climate Data, DEM, NDVI, Boundary Data
The data sources can be grouped into four categories. These categories are climate data, DEM data, NDVI data and state boundary data. The climate datasets are obtained from 75 different weather stations across NSW and ACT. The raw datasets can be obtained from the Australian Bureau of Meteorology (BOM) website \citep{bom2020datadir}. In the climate data category, there are two groups of datasets. The first group consists of datasets from the year 2017. These datasets are used for model training, validation, and preliminary testing. The second group is datasets from June, July, and August (winter \citep{bom2018winter}) of 2018. These datasets are used to test the model performance in the winter one year after the models are constructed. The winter data is used because the estimated potential frost days cover most of the study area during these winter months \citep{bomfrost}.

% DEM
The second data source category includes a DEM dataset of the study area. The DEM dataset contains 1 second (about 30 m) resolution elevation data of Australia. The NSW and ACT portion of the DEM dataset is extracted to be used as features during model construction, validation, and testing. The raw DEM dataset is hosted by Geoscience Australia \citep{STRMDEM}.

% NDVI
The raw NDVI datasets are collected from the Land Processes Distributed Active Archive Center of NASA \citep{MODISNDVI}. The data resolution is 250 m. Similar to the climate datasets, NDVI datasets are from the year 2017 and the winter months of 2018. The datasets from 2017 are also used for model training, validation, and preliminary testing. The NDVI datasets from 2018 act as features during the final testing phase to evaluate the model performance in the winter one year after model construction.

% Boundary Data
Boundary datasets form the final data category in this article. There are two separate data files that contain boundary information of NSW and ACT. The area coverage is defined by coordinates of multiple polygon vertices. The boundary datasets are not directly involved in the process of model training, validation, or testing. However, the boundary datasets are required when extracting the NSW and ACT portion of the DEM and NDVI datasets. The boundary datasets of NSW and ACT can be obtained from \citep{NSWBound} and \citep{ACTBound}, respectively.

\subsection{Data Preprocessing}
Data preprocessing procedures are described according to the data categories defined in the above subsection. Climate data preprocessing is explained first, followed by DEM data preprocessing and NDVI data preprocessing. The boundary data is used during the preprocessing of DEM and NDVI datasets. However, as boundary data is not included in the processes of model training, validation, and testing, the usage of boundary data is only presented with other data categories. After the preprocessing of the four data categories, the datasets are merged to be fit as training data of prediction models. This merging process is also explained.

% Climate
The temperature, dew point, Relative Humidity (RH), wind speed, and wind direction fields are extracted from each of the weather stations. Wind speed and direction are converted to \textit{N-wind} and \textit{E-wind}, as the north and east components of the wind. This conversion is done by reversing the wind direction ($met$) to the wind blowing direction ($deg$) (Equation \ref{eqn:temp_eqn_1}) \citep{6974024}. Then, calculating the magnitude of the eastward ($v_E$) and northward ($v_N$) wind components with the wind speed ($v$) and wind blowing direction ($deg$) (Equation \ref{eqn:temp_eqn_2}). After computing the wind components, the next hour minimum temperature is calculated as the training targets of the models. For each time step at each weather station, the prediction target is obtained from the minimum temperature of the next 60 time steps.

%\begin{equation}
%\label{eqn:temp_eqn_1}
%   deg= \begin{cases} 
%        met+180^\circ, & \text{if}\ met<180^\circ\\ 
%        met-180^\circ, & \text{if}\ met\geq 180^\circ \end{cases}
%\end{equation}
\begin{equation}
\label{eqn:temp_eqn_1}
   deg=met+180^\circ,\:if\:met<180^\circ;\:deg=met-180^\circ,\:if\:met\geq 180^\circ
\end{equation}

\begin{equation}
\label{eqn:temp_eqn_2}
    v_E,v_N=v\times sin(deg),v\times cos(deg)
\end{equation}

\begin{figure}[!htpb]
\centering
\includegraphics[width=6.4in]{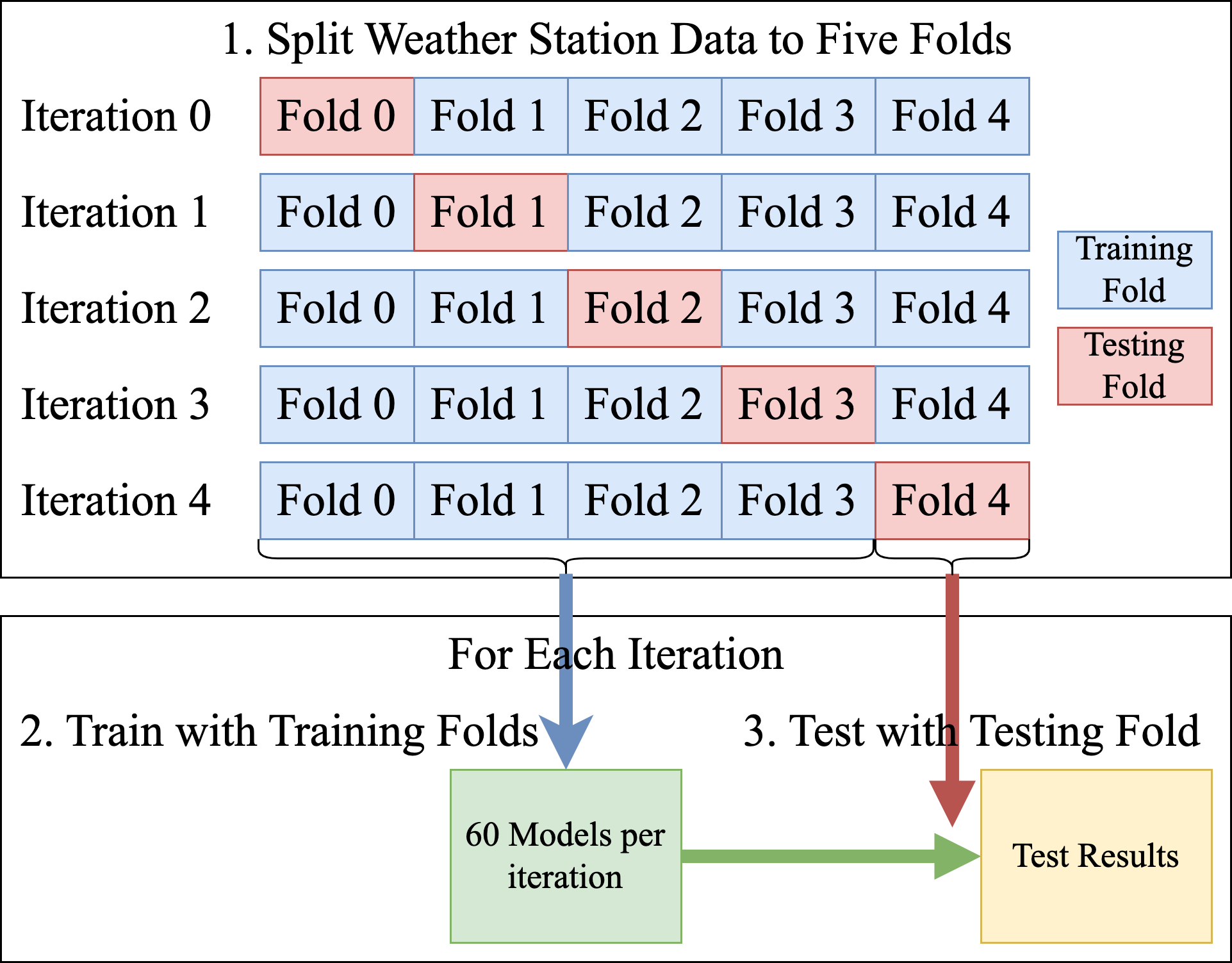}
\DeclareGraphicsExtensions{.pdf,.jpeg,.png,}
\caption{Training and Testing with Five-fold Validation.}
\label{fig:spat_folds}
\end{figure}

\begin{figure}[!htpb]
\centering
\includegraphics[width=6.4in]{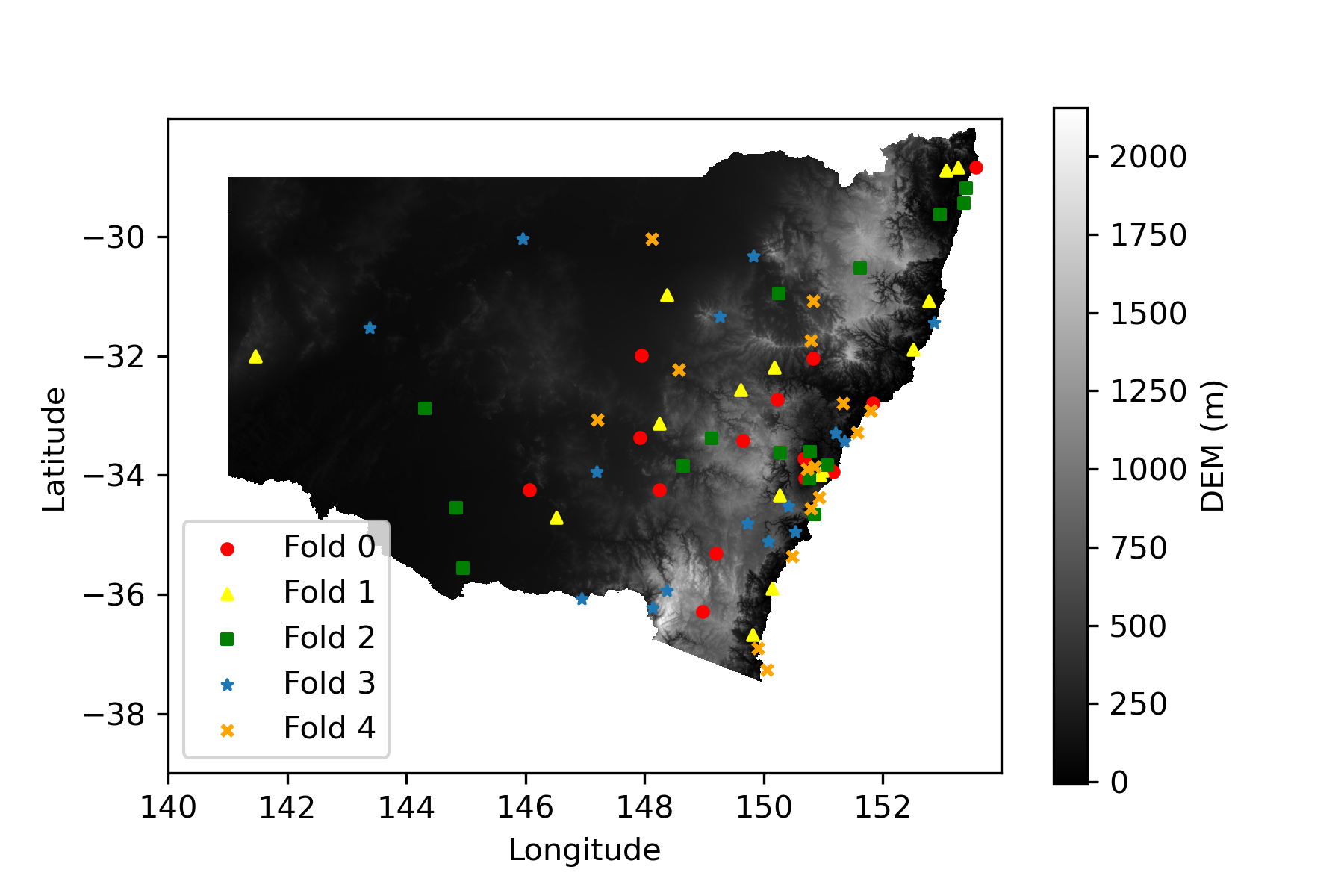}
\DeclareGraphicsExtensions{.pdf,.jpeg,.png,}
\caption{Distribution of Station Folds on NSW and ACT DEM Map.}
\label{fig:spat_stations}
\end{figure}

Since the spatial interpolation models are built with five-fold validation, the weather stations are randomly divided into five testing folds \citep{gareth2013introduction}. For each model fold, one different fold of the weather stations (15 stations) is used as testing data sources and the rest are used as training data sources. There are 60 submodels corresponding to the 60 stations. Each model uses one station as the “source station” to provide environmental readings or climate features. The other stations provide the labels or target (minimum temperature) of the training data entries. Each entry connects a source station to a target station. Therefore, each entry also contains the station attributes (longitude, latitude, DEM, NDVI) of the source and target stations. The testing stations are only involved during testing as target stations. Figure \ref{fig:spat_folds} summarizes the above steps of five-fold validation and figure \ref{fig:spat_stations} shows the distribution of weather station folds.

% DEM NDVI
The DEM dataset is resampled to a grid with a cell size of $0.01^{\circ}\times 0.01^{\circ}$ (approximately 1.11 km $\times$ 1.11 km). Then, the NSW and ACT parts of the map are extracted according to the boundary datasets. The DEM readings of Figure \ref{fig:spat_stations} is the result of DEM preprocessing. The NDVI datasets are preprocessed with the same procedures as the DEM dataset. These datasets are also resampled to a $0.01^{\circ}\times 0.01^{\circ}$ grid. Then, data is extracted within NSW and ACT boundaries.

% Merged
Table \ref{tab:spat_input} shows five features groups. The training data required by all models in this article can be represented by assembling some of the five feature groups. There are two data input formats that correspond to the two model types. The first model type is the baseline model. This model type requires on-site climate data for prediction. Therefore, the source station climate features are required as the input features of these models, and the next hour minimum temperature of the source station is used as the model target or predictand.

The second model type provides off-site frost predictions. For each model prediction, the models utilize climate data from a source station and predict the next hour minimum temperatures for a target station in another location. The source station attributes and target station attributes are acquired as the features to compute the differences between stations (Location, DEM, NDVI). Also, the source station climate features are input as climate references. Finally, the spatial model predictand is the minimum temperature values from the target station.

\begin{table}[!htbp]
%% increase table row spacing, adjust to taste
\renewcommand{\arraystretch}{1.3}
% if using array.sty, it might be a good idea to tweak the value of
%\extrarowheight as needed to properly center the text within the cells
\caption{Features and predictands of preprocessed datasets}
\label{tab:spat_input}
\centering
%% Some packages, such as MDW tools, offer better commands for making tables
%% than the plain LaTeX2e tabular which is used here.
\begin{tabular}{| p{\dimexpr 0.37\linewidth-2\tabcolsep}| p{\dimexpr 0.55\linewidth-2\tabcolsep}|}
\hline
\textbf{Feature Groups} & \textbf{Features}\\
\hline
Source Station Attributes&Longitude, Latitude, DEM, NDVI\\\hline
Target Station Attributes&Longitude, Latitude, DEM, NDVI\\\hline
Source Station Climate Features&Temperature, Dew Point, RH, N-wind, E-wind\\\hline
Baseline Predictand&Source Station Minimum Temperature\\\hline
Spatial Model Predictand&Target Station Minimum Temperature
\\\hline
\end{tabular}
\end{table}

\subsection{Model Types}
% Baseline Model
This article involves two major model types. They are the baseline models and spatial interpolation-based models. Baseline models are ANN models that harness data from sensors at a particular site and predict frost on the site. Every model is designed to predict frost for one different location. These models represent recent frost prediction models that are built from on-site historical climate datasets. The baseline models are three-layer fully connected models. Starting from the first layer, the number of nodes is five, seven and one. The first two layers utilize the rectifier activation function, and the final layer is linear.

% Spatial Interpolation Model
Spatial interpolation-based models are also ANN models. Unlike the baseline models, the spatial interpolation-based models do not need on-site sensor data from the target location. These models use off-site climate data from weather stations of other locations to predict any frost of the target location. In this article, there are multiple spatial interpolation-based models. Each model in each fold is a collection of multiple submodels or weak predictors. Each submodel is constructed based on the climate data of a corresponding station as the source station, and the other stations (target stations) from the training stations provide the minimum temperature labels. The submodels are five-layer fully connected models. Starting from the first layer, the number of nodes is 10, 14, nine, eight and one. The first four layers utilize the rectifier activation function, and the final layer is linear. To increase the accuracy of predictions, the results of all submodels within a fold are aggregated. Three results aggregation methods are tested. The first method is averaging all results within each fold. The second method is to compute a weighted average based on the difference between each source station and the target location/station. This difference is calculated by geographical coordinates, DEM, and NDVI. The final method or the weighted voting method does not aggregate the predicted minimum temperature. Each submodel provides a weighted vote of whether there is a frost event (+1) or not (-1) based on the predicted minimum temperature. The aggregated vote is positive when the prediction result is smaller than the triggering temperature zero degrees, else the vote is negative. The weights of the weighted vote method are obtained by using the same algorithm as the weighted average method.

% Weighted Average Layer
\subsubsection{Weighted Average Layer}
The weighted average layer is applied after obtaining the results of multiple spatial interpolation-based models in a particular fold. A weight is calculated for every station with an available prediction result. An intermediate weight for the \textit{i}th source station ($W_i$) is first calculated by Equation \ref{eqn:spat_eqn_3}. $g_i$, $d_i$, and $n_i$ are normalized geographical distance, DEM difference, and NDVI difference between the source station and the target location. The geographical distance is computed as a two-dimensional Euclidean distance. DEM and NDVI differences are one-dimensional Euclidean distances. The coefficients $a$, $b$, and $c$ provide adjustable importance of the geological distance, DEM difference, and NDVI difference to the station weight. Currently, the adjustable weights (Table \ref{tab:spat_weight}) are generated by Pearson Correlation between each of the distance/differences and model errors. After the intermediate weights for all stations are obtained, these intermediate weight for each station is divided by the sum of the intermediate weights. The final station weights are obtained by normalizing the divided intermediate weights.

\begin{equation}
\label{eqn:spat_eqn_3}
W_i=\frac{1}{ag_i+bd_i+cn_i}
\end{equation}

\begin{table}[!htbp]
%% increase table row spacing, adjust to taste
\renewcommand{\arraystretch}{1.3}
% if using array.sty, it might be a good idea to tweak the value of
%\extrarowheight as needed to properly center the text within the cells
\caption{Adjustable Weights for Different Folds}
\label{tab:spat_weight}
\centering
%% Some packages, such as MDW tools, offer better commands for making tables
%% than the plain LaTeX2e tabular which is used here.
\begin{tabular}{|p{\dimexpr 0.23\linewidth-2\tabcolsep}| p{\dimexpr 0.21\linewidth-2\tabcolsep} | p{\dimexpr 0.23\linewidth-2\tabcolsep}|p{\dimexpr 0.24\linewidth-2\tabcolsep}|}
\hline
\textbf{Fold Number} & \textbf{Geo Weight ($a$)} & \textbf{DEM Weight ($b$)} & \textbf{NDVI Weight ($c$)}\\
\hline
0&0.1629& 0.0132& 0.0290\\\hline
1&0.1768& 0.0205& 0.0238\\\hline
2&0.1612& 0.0222& 0.0177\\\hline
3&0.1804& 0.0114& 0.0269\\\hline
4&0.1601& 0.0110& 0.0260
\\\hline
\end{tabular}
\end{table}

\subsection{Experiments}
To compare the results of the off-site prediction models with the baseline, three experiments are conducted. The aim of the first experiment is to compare the prediction results of different folds and determine if bias exists due to different station training data. Three sets of raster maps of NSW and ACT are generated from the datasets acquired in the year 2017. These raster maps are computed from the spatial interpolation-based models. Each map of the first set is created using the results from one different weather station. The second set is created by averaging the first map set per fold. After that, the final map set is generated using the weighted average layer.

The second experiment compares the accuracy between the three model types (baseline, averaging, weighted averaging). The testing datasets are from 2017. The baseline models use parts of the datasets not involved in the training process (20\% of data for each station from 2017). For spatial interpolation-based models, 15 weather stations per fold act as testing datasets.

In the final experiment, datasets from the winter months of 2018 are used to compare the spatial interpolation-based models with different numbers of available stations. Model accuracy is compared against the baseline models. After that, the percentages of captured events below zero degrees \citep{snyder2005frost} are also evaluated. An event is determined as a time step, where the temperature is below zero degrees. An event is captured when the prediction algorithm predicts that the temperature is below zero degrees at the time step of the event. Finally, the results of the proposed methods are compared with the results of Inverse Distance Weighting (IDW) and Ordinary Kriging (OK). IDW and OK are the two most commonly compared spatial interpolation methods \citep{LI2014173}.

\section{Results and Discussions}
\label{spat:result}
% Comparison between raster maps: individual station, averaging, weighted averaging
% Comparison between accuracy: baseline, individual station, averaging, weighted averaging.
% Comparison between accuracy: Next Year (June, July, August)
% Comparison between percentage of captured events below zero degrees: Next Year (June, July, August)

\subsection{Effect of Different Fold Training Datasets}
% 63291,66137,58212,72160,67119

In this experiment, the differences in the results generated by models trained and tested with different datasets are revealed. As examples, one test station is randomly chosen from each of the weather station folds. Each test station is also a source station in the other four folds. Therefore, using each test station as a source station, four raster maps are generated corresponding to the other four folds. These raster maps are generated from four different models using the same climate data, but different prediction targets during training. The five chosen stations are stations 63291, 66137, 58212, 72160, and 67119, from folds 0, 1, 2, 3, and 4, respectively.

\begin{figure}[!htpb]
\centering
\includegraphics[width=6.4in]{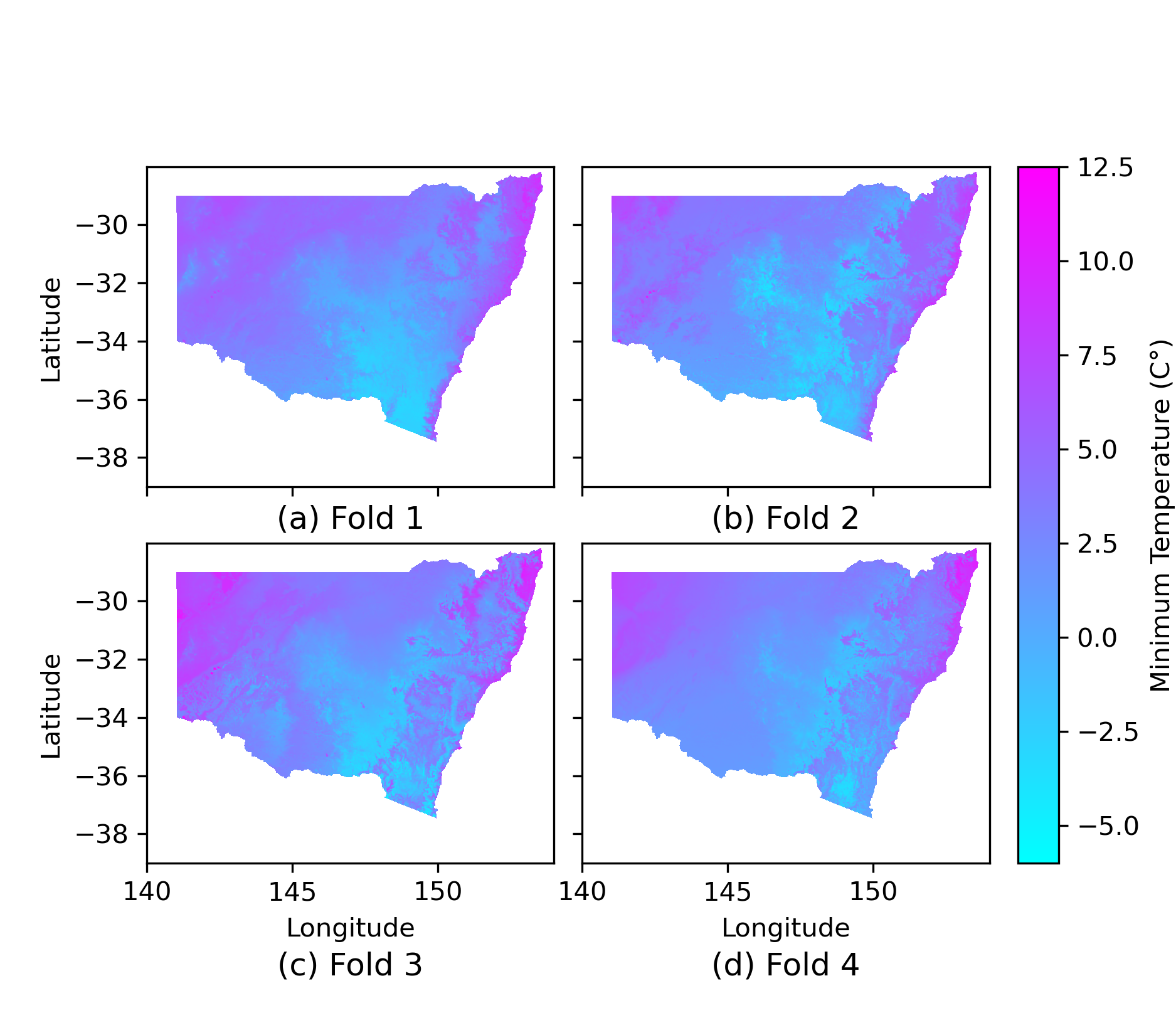}
\DeclareGraphicsExtensions{.pdf,.jpeg,.png,}
\caption{Raster Maps from Models Using Weather Station 63291 as Climate Data Source Trained with the Datasets from Folds 1--4.}
\label{fig:spat_raster_63297}
\end{figure}

\begin{table}[!htbp]
%% increase table row spacing, adjust to taste
\renewcommand{\arraystretch}{1.3}
% if using array.sty, it might be a good idea to tweak the value of
%\extrarowheight as needed to properly center the text within the cells
\caption{P-value Matrix for Comparing Raster Map Results of Models Using Weather Station 63291 as Climate Data Source and Trained with the Datasets from Folds 1--4}
\label{tab:spat_63297}
\centering
%% Some packages, such as MDW tools, offer better commands for making tables
%% than the plain LaTeX2e tabular which is used here.
\begin{tabular}{|p{\dimexpr 0.18\linewidth-2\tabcolsep}| p{\dimexpr 0.18\linewidth-2\tabcolsep} | p{\dimexpr 0.18\linewidth-2\tabcolsep}|p{\dimexpr 0.18\linewidth-2\tabcolsep}|p{\dimexpr 0.18\linewidth-2\tabcolsep}|}
\hline
& \textbf{Fold 1} & \textbf{Fold 2} & \textbf{Fold 3} & \textbf{Fold 4}\\
\hline
\textbf{Fold 1}&N/A& 0.00& 7.87e-61&0.00\\\hline
\textbf{Fold 2}&0.00& N/A& 0.00&0.00\\\hline
\textbf{Fold 3}&7.87e-61& 0.00& N/A&0.00\\\hline
\textbf{Fold 4}&0.00& 0.00& 0.00&N/A
\\\hline
\end{tabular}
\end{table}

Figure \ref{fig:spat_raster_63297} shows the raster maps generated from individual models of station 63291 constructed by the training data (excluding the data from testing stations in each test fold) in fold 1--4. From a visual perspective, the raster maps of the four folds are similar in major features. For example, all maps present heat spots in the northeastern and northwestern regions. Also, all maps demonstrate the cold spot near the center of the maps. However, the shapes of these features are different between each fold. Folds 2 and 3 show more details than the other two folds. The maps seem to be significantly affected by the available training datasets for different folds. Table \ref{tab:spat_63297} also supports the hypothesis that raster map results are affected by the different training datasets in each fold. Table \ref{tab:spat_63297} is the P-value matrix by conducting paired T-tests for all possible fold raster map result pairs. All of the P-values are smaller than 0.05. This rejects the null hypothesis and favors the alternative hypothesis that the raster maps are different from each other. Raster maps from stations 66137, 58212, 72160, 67119 draw similar conclusions. The raster maps of these stations are placed in \ref{appx:spat_compare} and the P-value matrices are placed in \ref{appx:spat_ttest}.

\begin{figure}[!htpb]
\centering
\includegraphics[width=6.4in]{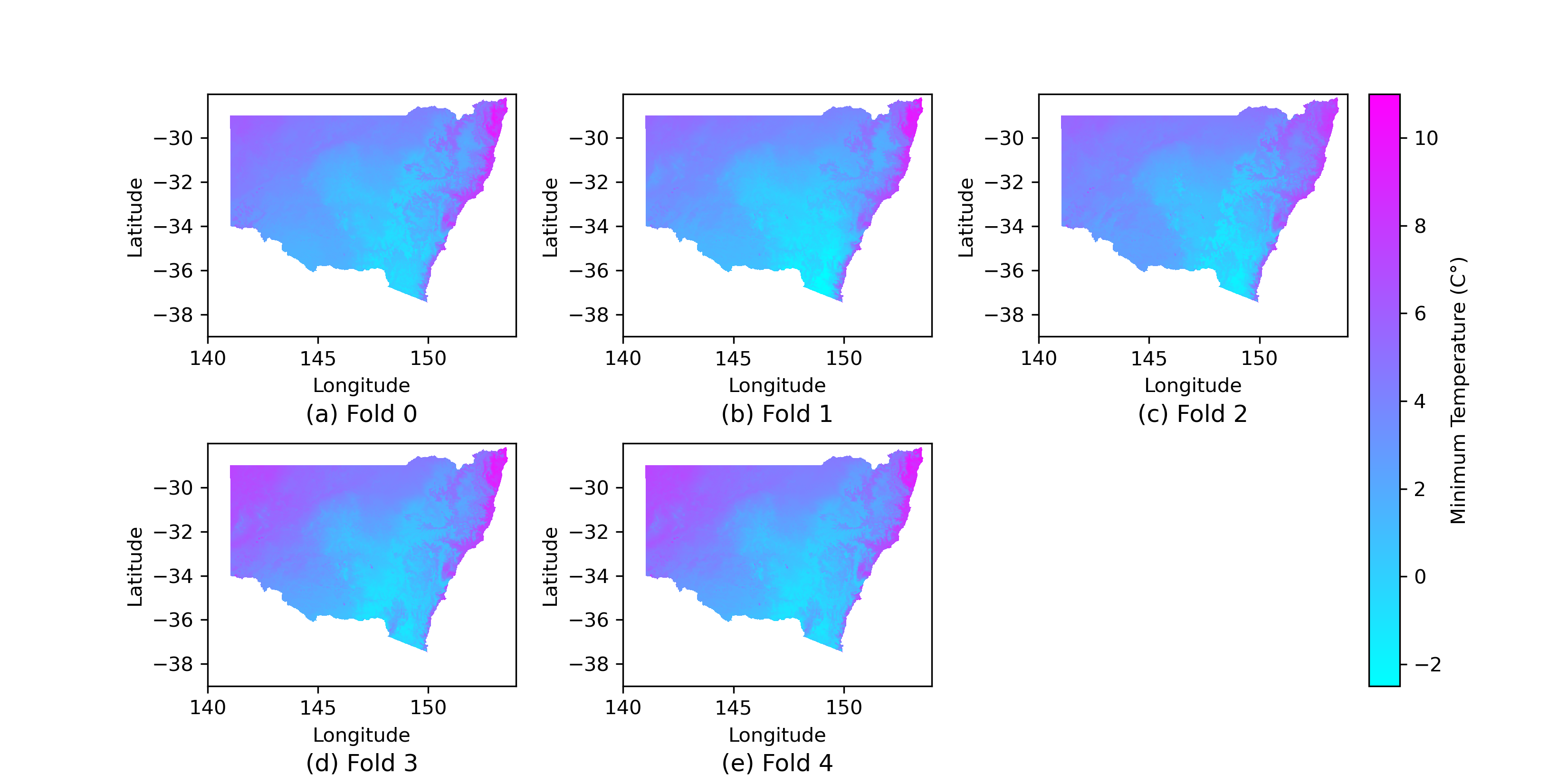}
\DeclareGraphicsExtensions{.pdf,.jpeg,.png,}
\caption{Raster Maps from Averaged Results per Fold.}
\label{fig:spat_raster_average}
\end{figure}

Raster maps (Figure \ref{fig:spat_raster_average}) are also generated for the averaged result of each folds. Compared to the maps generated from individual models, maps of averaging models per fold are more visually similar. However, The results of paired T-tests (Table \ref{tab:spat_average}) are similar to the individual models. All of the P-values are smaller than 0.05. This suggests strong evidence against the null hypothesis. There are significant differences between these averaged maps from different folds. These differences also exist between weighted averaged maps.

\begin{table}[!htbp]
%% increase table row spacing, adjust to taste
\renewcommand{\arraystretch}{1.3}
% if using array.sty, it might be a good idea to tweak the value of
%\extrarowheight as needed to properly center the text within the cells
\caption{P-value Matrix for Comparing Averaged Raster Map Results of Each Fold}
\label{tab:spat_average}
\centering
%% Some packages, such as MDW tools, offer better commands for making tables
%% than the plain LaTeX2e tabular which is used here.
\begin{tabular}{|p{\dimexpr 0.14\linewidth-2\tabcolsep}| p{\dimexpr 0.14\linewidth-2\tabcolsep} | p{\dimexpr 0.14\linewidth-2\tabcolsep}|p{\dimexpr 0.18\linewidth-2\tabcolsep}|p{\dimexpr 0.14\linewidth-2\tabcolsep}|p{\dimexpr 0.18\linewidth-2\tabcolsep}|}
\hline
& \textbf{Fold 0} & \textbf{Fold 1} & \textbf{Fold 2} & \textbf{Fold 3}& \textbf{Fold 4}\\
\hline
\textbf{Fold 0}&N/A&0.00&0.00&0.00&0.00\\\hline
\textbf{Fold 1}&0.00& N/A& 0.00&0.00&0.00\\\hline
\textbf{Fold 2}&0.00& 0.00& N/A&0.00&2.67e-121\\\hline
\textbf{Fold 3}&0.00& 0.00& 0.00&N/A&0.00\\\hline
\textbf{Fold 4}&0.00& 0.00& 2.67e-121&0.00&N/A
\\\hline
\end{tabular}
\end{table}

Finally, raster maps (Figure \ref{fig:spat_raster_weighted} in \ref{appx:spat_compare}) created from weighted averages of models within each fold present similar properties to the averaged maps. From the T-tests (Table \ref{tab:spat_weighted} in \ref{appx:spat_ttest}), the effect of different training station datasets still persists.

\subsection{Model Accuracy}

In the second experiment, model accuracy is compared between individual station models for different folds, an averaged station result per fold, a weighted average result of station models per fold, and the baseline. The baseline, using on-site sensor data, reaches the highest accuracy with the lowest Root Mean Square Error (RMSE). The ensemble methods of averaging and weighted averaging outperform the individual station models in the spatial interpolation-based methods. Ensemble by weighted average reached higher accuracy because higher weights are given to stations that are much similar to the target locations.

\begin{figure}[!htpb]
\centering
\includegraphics[width=6.4in]{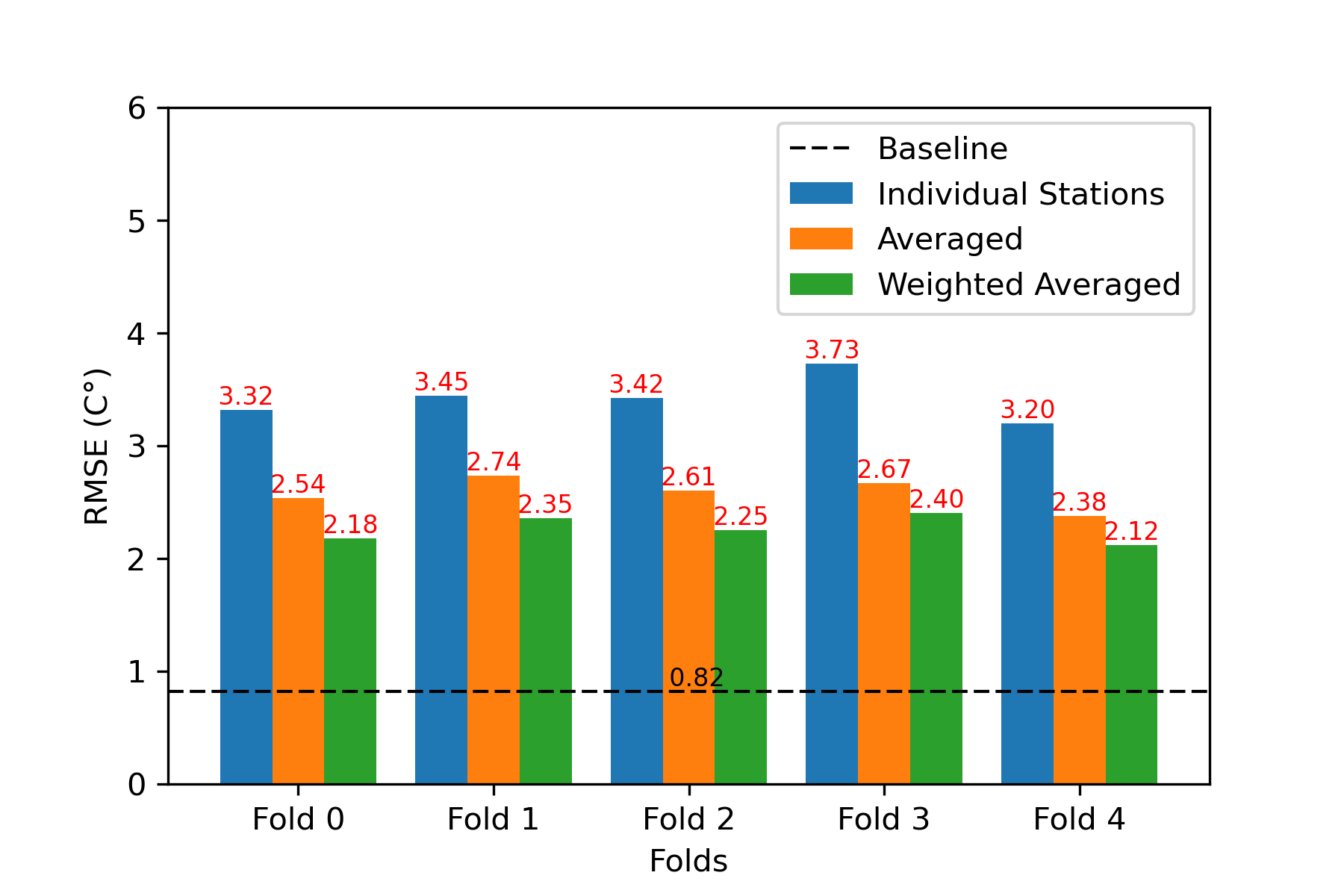}
\DeclareGraphicsExtensions{.pdf,.jpeg,.png,}
\caption{RMSE of Individual, Averaged, Weighted Averaged Models Obtained from Year 2017 Testing Datasets.}
\label{fig:spat_raster_accuracy}
\end{figure}

\subsection{Effect of the Number of Available Weather Stations}

In the final experiment, the ability of fault tolerance of the proposed ensemble methods is tested. The accuracy and event capture rate of the averaged and weighted averaged methods are tested with different numbers of available weather stations. This experiment focuses more on the ability to capture potential events below zero degrees in the future. Most of the baseline models achieved 100\% of event detection rate (true positive rate) in this experiment. The mean true positive rate of the baseline models is 99.62\%. To ensure fairness of the experiment, the testing datasets are from the year 2018 (a year after the training datasets).

\begin{figure}[!htpb]
\centering
\includegraphics[width=6.4in]{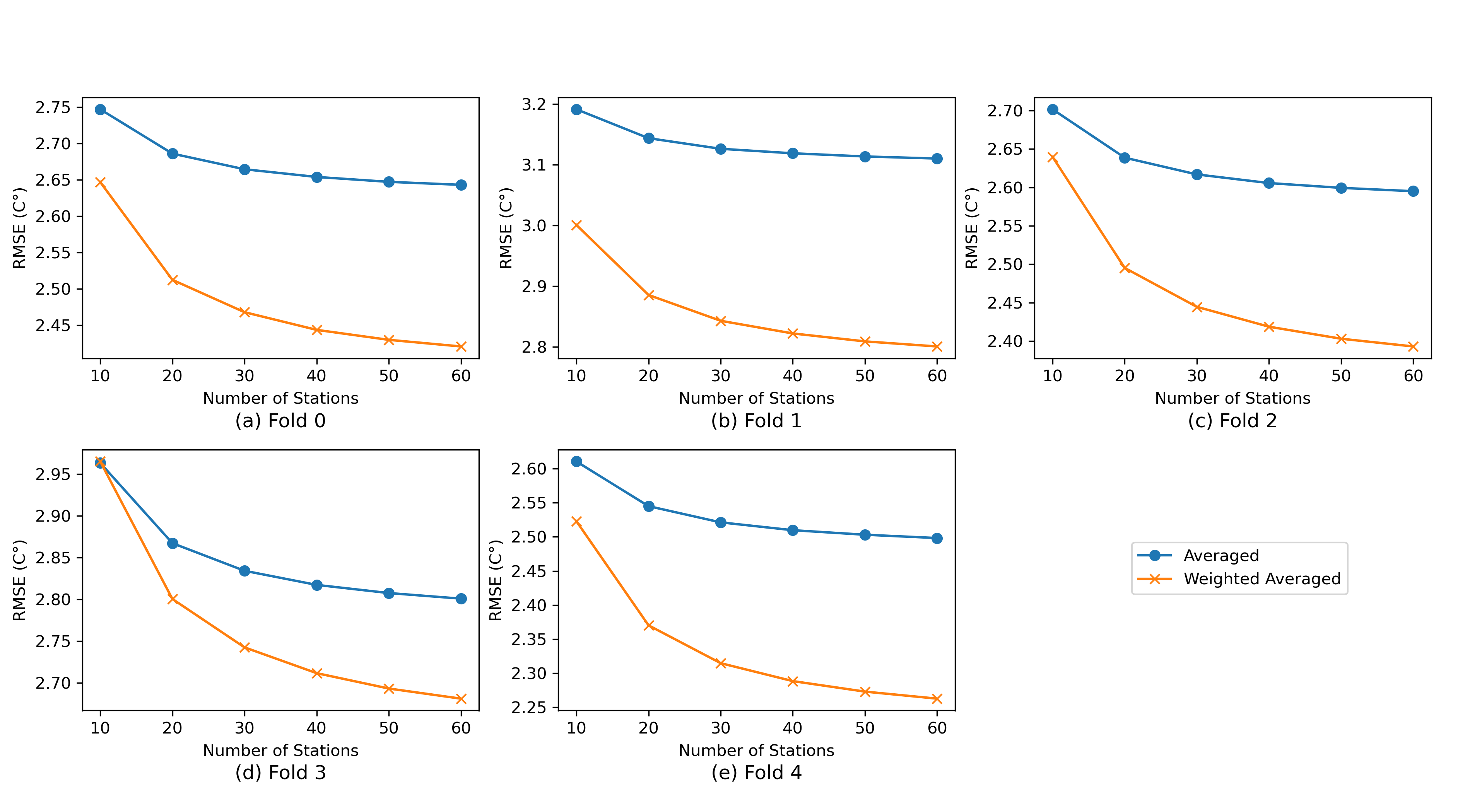}
\DeclareGraphicsExtensions{.pdf,.jpeg,.png,}
\caption{RMSE of Averaged and Weighted Averaged Models with 10--60 Available Stations Obtained from Year 2018 Testing Datasets.}
\label{fig:spat_raster_1060RMSE}
\end{figure}

Figure \ref{fig:spat_raster_1060RMSE} shows the accuracy of the proposed ensemble methods with different numbers of available weather stations 10, 20, 30, 40, 50, and 60. The weighted averaged method outperforms the averaged method with a lower RMSE. Accuracy for both methods increases as the number of stations increases.

\begin{figure}[!htpb]
\centering
\includegraphics[width=6.4in]{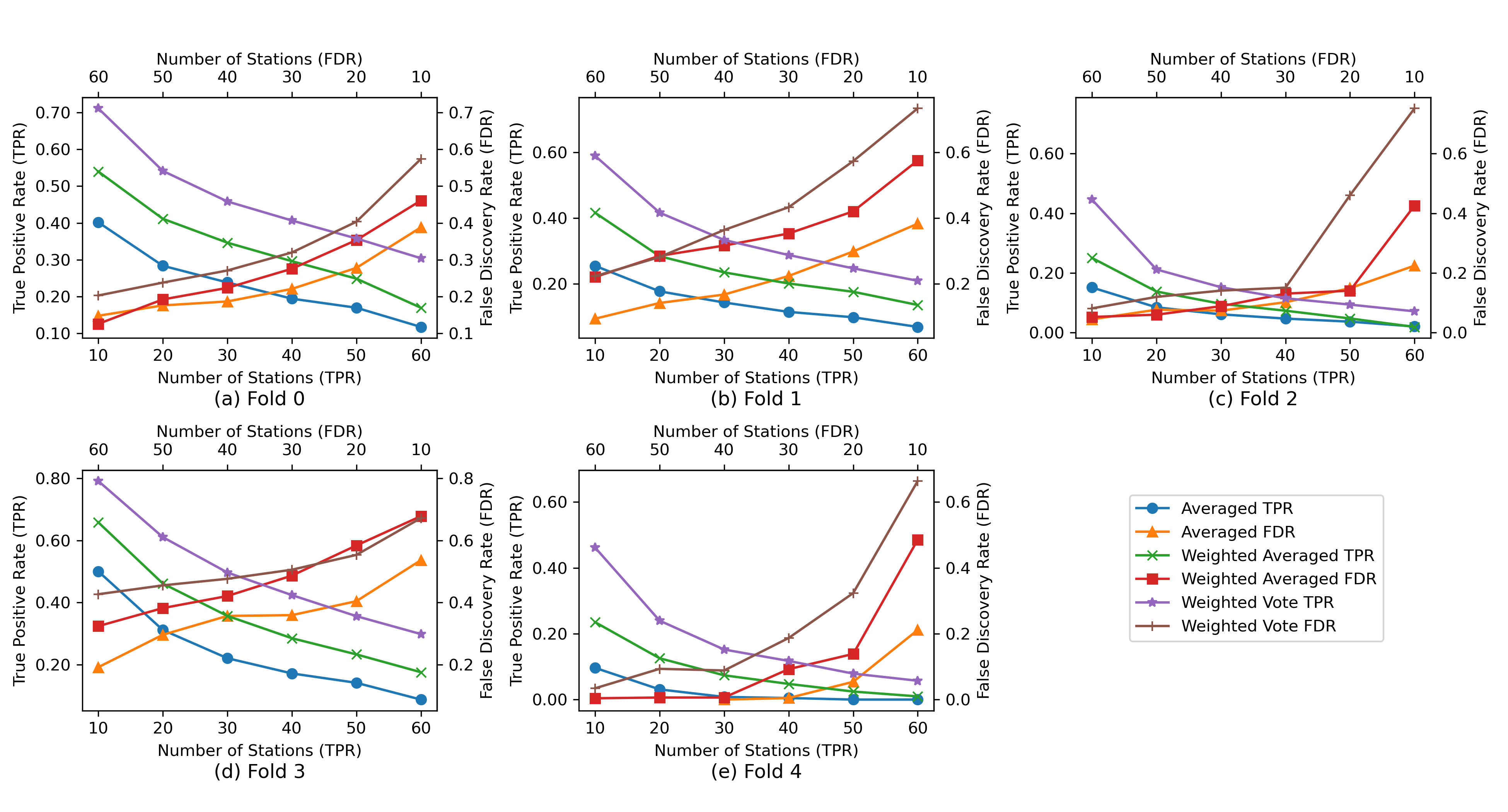}
\DeclareGraphicsExtensions{.pdf,.jpeg,.png,}
\caption{True Positive Rate and False Discovery Rate of Averaged, Weighted Averaged, and Weighted Voting Models with 10--60 Available Stations Obtained from Year 2018 Testing Datasets.}
\label{fig:spat_raster_1060TPRFDR}
\end{figure}

Contrary to the accuracy, the event capture rate or true positive rate reduces as the number of stations increases (Figure \ref{fig:spat_raster_1060TPRFDR}). However, the false discovery rate also reduces with the increase of the number of available stations. This illustrates that the event detection rate is high due to high false positives. The ensemble of results with more stations reduces error. However, high errors induce lower temperature predictions, which increase the event capture rate with the number of false positives. Another experiment with the number of available stations 1--10 is conducted to inspect the relationship between true positive rate and false discovery rate.

\begin{figure}[!htpb]
\centering
\includegraphics[width=6.4in]{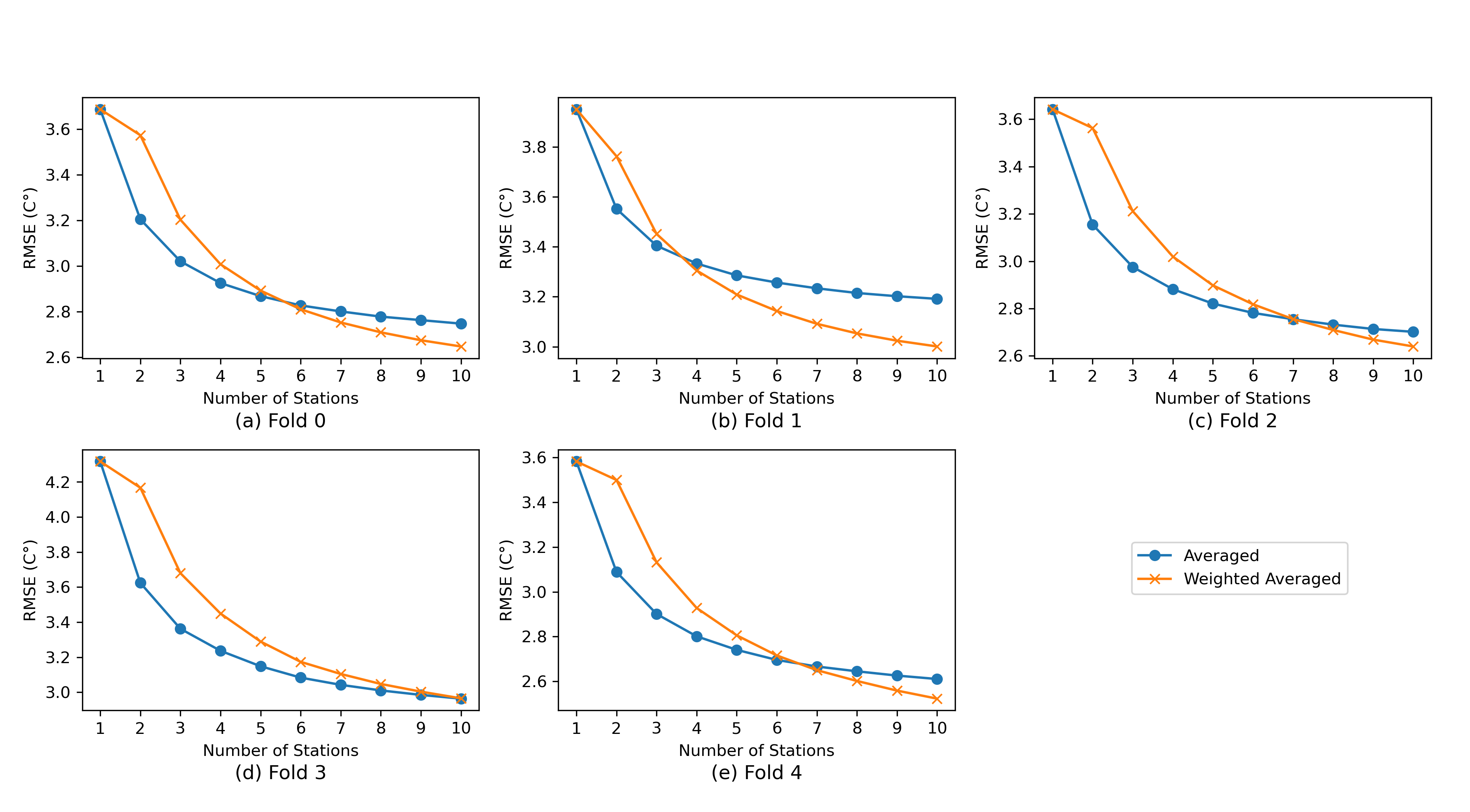}
\DeclareGraphicsExtensions{.pdf,.jpeg,.png,}
\caption{RMSE of Averaged and Weighted Averaged Models with 1--10 Available Stations Obtained from Year 2018 Testing Datasets.}
\label{fig:spat_raster_0110RMSE}
\end{figure}

\begin{figure}[!htpb]
\centering
\includegraphics[width=6.4in]{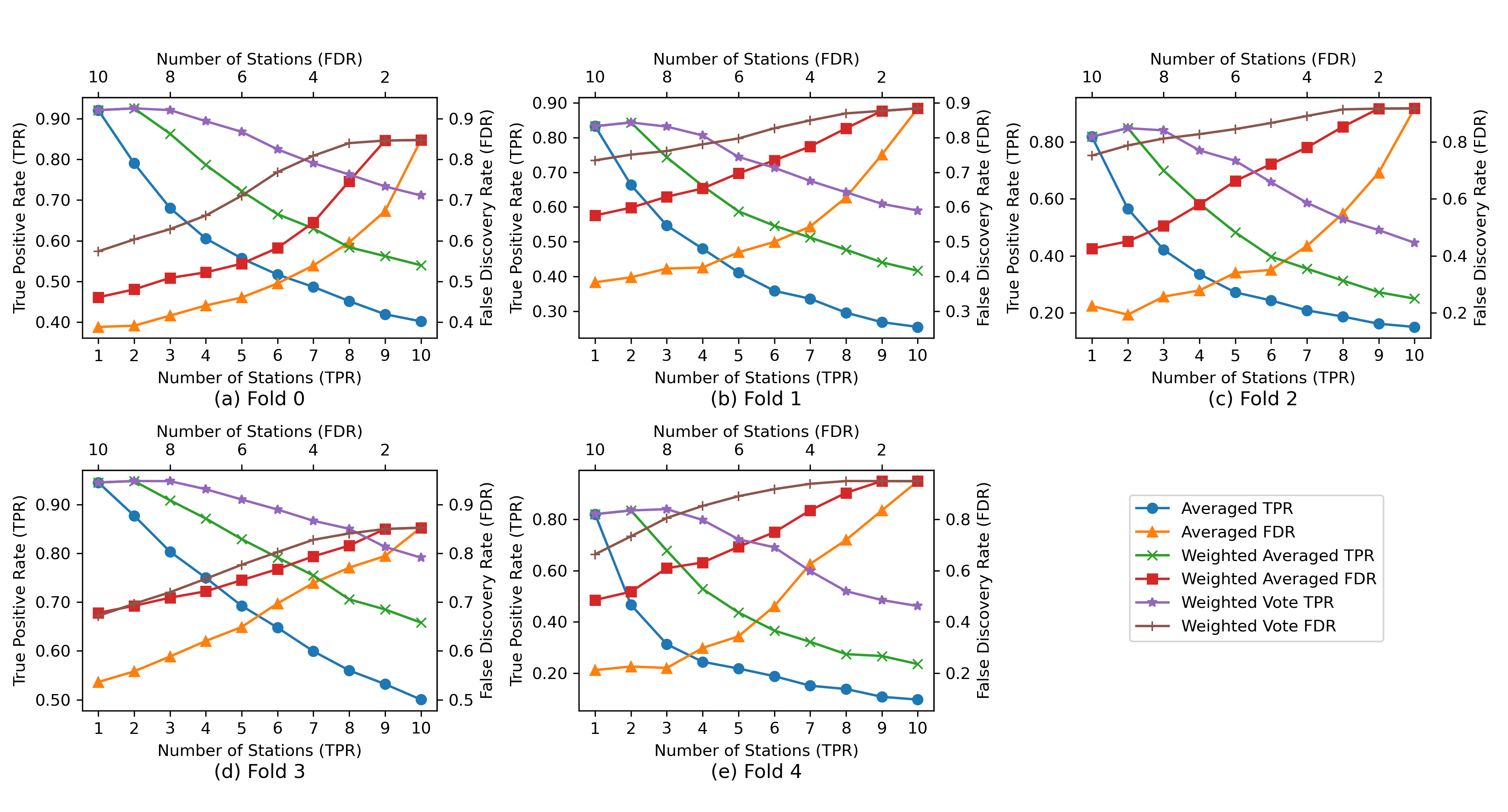}
\DeclareGraphicsExtensions{.pdf,.jpeg,.png,}
\caption{True Positive Rate and False Discovery Rate of Averaged, Weighted Averaged, and Weighted Voting Models with 1--10 Available Stations Obtained from Year 2018 Testing Datasets.}
\label{fig:spat_raster_0110TPRFDR}
\end{figure}

The RMSE (Figure \ref{fig:spat_raster_0110RMSE}) of the ensemble methods for 1--10 stations follows previous patterns. RMSE reduces with the increase of the number of available stations. The true positive rate performs a more extreme pattern. When there is only one available station, the true positive rate could exceed 90\% (92.55\% as the highest) (Figure \ref{fig:spat_raster_0110TPRFDR}). However, the false discovery rate also exceeds 90\% when there is only one available station. In conclusion, as the accuracy of temperature prediction reduces, the models recognize more events. This increase of event recognition increases the true positive rate along with the false discovery rate. The number of stations selected from the ensemble methods should be considering the balance between the event capture rate and number of false positives.

% Weighted voting method
Generally, the weighted averaged method demonstrates a higher accuracy and event detection rate compared to the averaged method. However, for both methods the true positive rate decreases as the number of stations increases. This phenomenon is due to the low accuracy of the weak predictors and the smoothing effect of the averaging methods. The averaging methods are filtering out the lower extremes of the prediction results. Therefore, a third method or the weighted voting method is proposed.

Figures \ref{fig:spat_raster_1060TPRFDR} and \ref{fig:spat_raster_0110TPRFDR} compared the true positive rate and false discovery rate of the weighted voting method with the previous two methods. The true positive rates have significantly increased compared to the previous methods, when utilizing the same number of stations. However, the false discovery rate also increased. To further increase the true positive rate without the increase of false discovery rate of the spatial interpolation-based methods, the accuracy of the weak predictor models should be further improved.

\subsection{Comparing Proposed Data Aggregation Methods with Traditional Methods}

\begin{figure}[!htpb]
\centering
\includegraphics[width=6.4in]{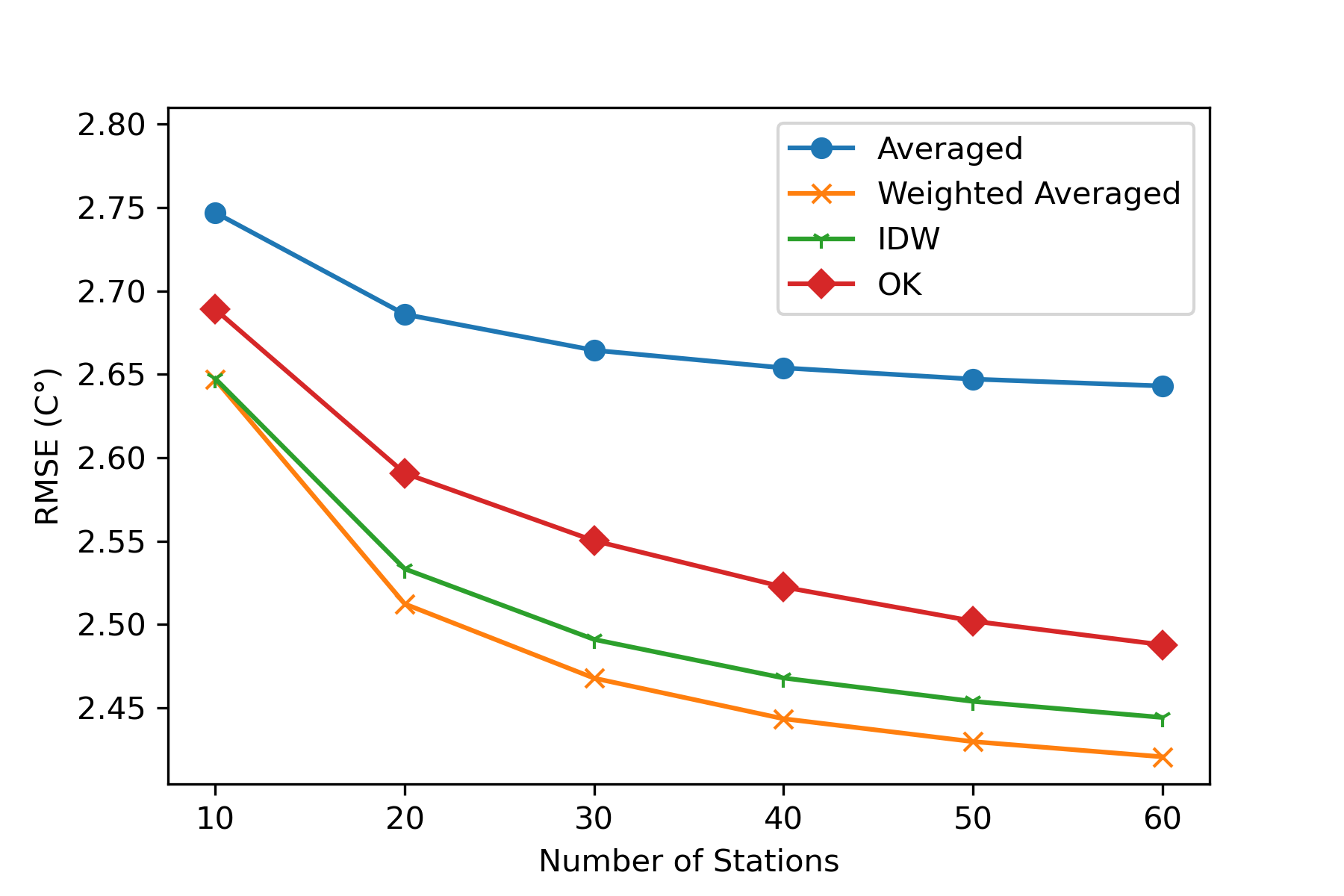}
\DeclareGraphicsExtensions{.pdf,.jpeg,.png,}
\caption{RMSE of Averaged, Weighted Averaged, IDW, and OK Models on 2018 Fold 0 Data.}
\label{fig:spat_extra_rmse}
\end{figure}

In this subsection, the proposed averaging, weighted averaging, and weighted voting methods are compared with two traditional kriging methods. At each time step, the prediction results of the 60 source weather stations in a data fold are aggregated by IDW and OK to compute the next hour minimum temperatures at the 15 testing weather stations. Similar to the previous subsection, the number of source stations is variable. The results presented are from Fold 0. From Figure \ref{fig:spat_extra_rmse}, the RMSEs of both IDW and OK are lower than the averaging method, but higher than the weighted averaging method. The traditional methods demonstrate similar patterns to the proposed methods in that a higher number of source stations reduces the RMSE.

\begin{figure}[!htpb]
\centering
\includegraphics[width=6.4in]{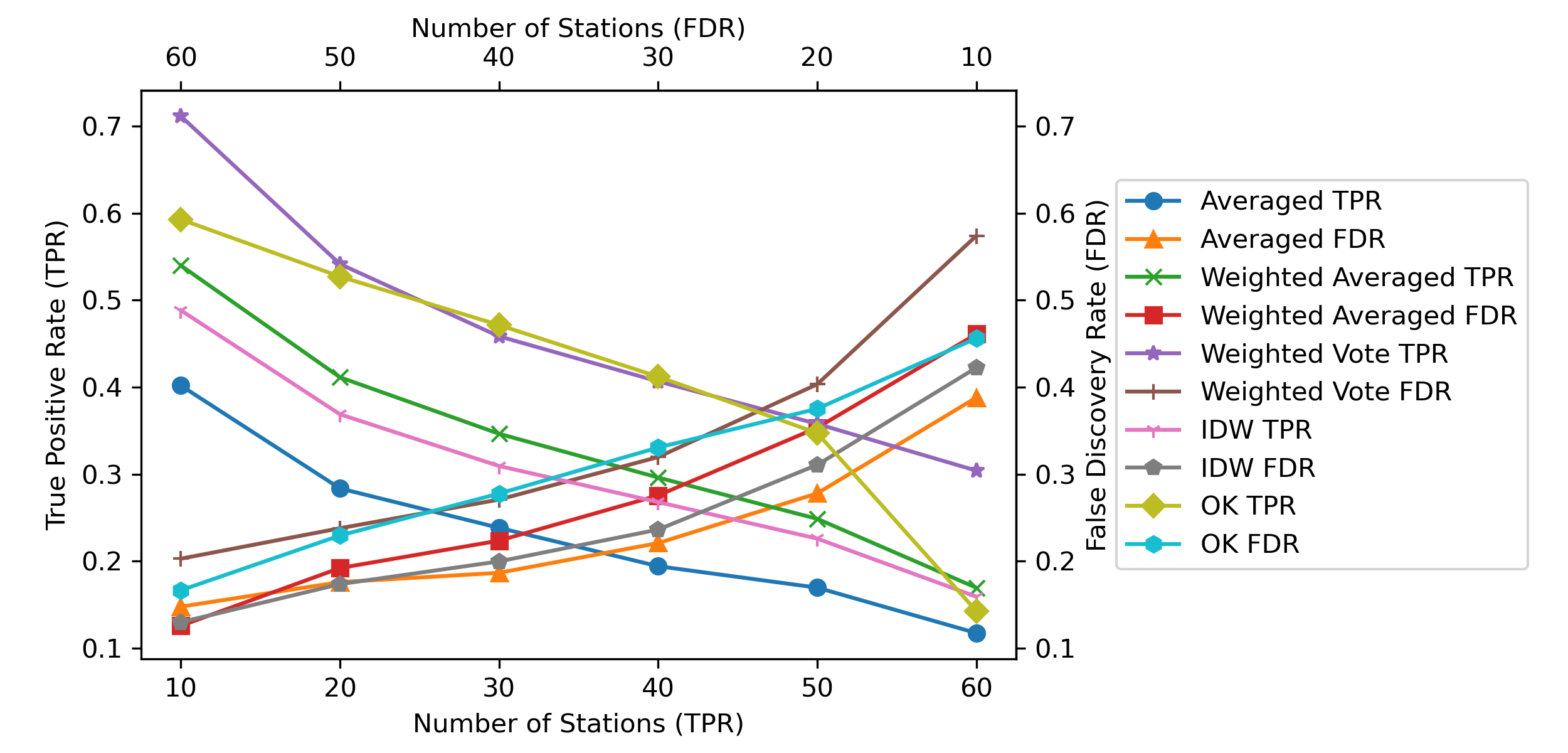}
\DeclareGraphicsExtensions{.pdf,.jpeg,.png,}
\caption{True Positive Rate and False Discovery Rate of Averaged, Weighted Averaged, Weighted Voting, IDW, and OK Models on 2018 Fold 0 Data.}
\label{fig:spat_extra_cap}
\end{figure}

Similarly, the true positive and false discovery rate trends of IDW and OK are also similar to the proposed methods (Figure \ref{fig:spat_extra_cap}). The true positive rates of the traditional methods are decreasing with the increase of source stations. The false discovery rates are decreasing with the increase of source stations. This shows that the low accuracy of the submodels also impacts the traditional methods. The true positive rates of the IDW method are only higher than the averaging method. In exchange, the false discovery rates of IDW are only higher than the averaging method. The true positive rates of OK are close to the weighted voting method. When there are 10 and 60 source stations, the true positive rates of the weighted voting method outperform OK. In exchange, the false discovery rates of OK at 10 and 60 stations are lower than the weighted voting method. In general, most of the proposed methods demonstrated a better ability to capture frost events than IDW. However, IDW demonstrated a lower false discovery rate. On the other hand, OK outperformed most of the proposed methods with a higher true positive rate except for the weighted voting method. However, OK also has a relatively high false discovery rate. Overall, the traditional methods still have the potential to be used as data aggregation methods for the submodels.

\subsection{Limitations and Open Challenges}
After testing the proposed method with the baseline, there are a few limitations of the method detected. These limitations are open challenges that lead toward future development of spatial interpolation-based frost prediction methods.

\subsubsection{Further improvement of the accuracy}
The errors of the proposed method are higher than the baseline created from on-site datasets. High event capture rates of the proposed method are only achieved with high numbers of false positives. To decouple event capture rate with the number of false positives, the increase of accuracy is inevitable. With the current accuracy, spatial interpolation-based methods cannot fully replace previous methods based on on-site historical datasets and sensors.

\subsubsection{Lacking of ground truth to validate models}
As spatial interpolation-based models are constructed with five-fold validation, the testing datasets are limited \citep{https://doi.org/10.1111/ecog.02881}. In this article, the testing datasets are generated from 15 weather stations for each fold. Therefore, any accuracy metrics can only represent the accuracy on these 15 weather stations. In the production environment, predictions are required on other locations. The model accuracy on other locations is uncertain. In future works, more weather stations could be involved to reduce this uncertainty.

\subsubsection{Models are highly sensitive to the choice of training weather stations}
As shown by experiment 1, raster maps produced from models of different folds are significantly different from each other. The models are affected by choices of training weather stations. From experiment 2, this choice also affects the accuracy of models. In future model construction with cross-validation, the accuracy of each fold should be carefully examined. A possible method to mitigate this issue is the increase in the number of weather stations. As the number of stations increases, the effect of individual stations reduces.

\subsubsection{Models can only predict for the next hour}
Due to the design of the proposed model, frost condition and minimum temperature can only be predicted for the next hour. Next hour predictions might be suitable for near realtime applications. However, it is still a limitation to obtaining future insights. Frost detection of the next few hours could be achieved by predicting future environmental data generated by time series predictions, such as \citep{CURCEAC2019394}. \citet{XIAO2019104502} have shown a spatiotemporal method to predict future environmental conditions with satellite data.

\subsubsection{Implementation with other spatial interpolation methods}
The focus of this paper is to eliminate the dependency on local sensors and data in frost prediction applications. Therefore, the proposed method is developed from previous machine learning-based frost prediction methods. However, there are also other factors that are not considered. The first factor is model design. Each submodel of the proposed method is an end-to-end model that directly produces the output minimum temperature of a target location. Could this be designed into a multi-stage pipeline model? For example, spatial interpolation is applied to the environmental parameters of a source station to obtain the parameters at the target location. Then, these environmental parameters could be fed to a local sensor/data-based prediction model to obtain the minimum temperature. These parameters could be reused for other applications at the target location. Another unconsidered factor is the usage of other existing spatial interpolation methods. In this paper, the test results of OK demonstrated some potential. There are other kriging methods. These methods are reviewed in \citep{LI20111647}. \citet{LI20111647} also stated other machine learning models used in spatial interpolation methods. These models are support vector machine, random forest, and neuro-fuzzy network. Other than these, there are also state-space models based on Bayesian inference \citep{bakar2015spatiodynamic, bakar2016hierarchical}.

\subsubsection{Application of similar methods to other domains}
The recent frost prediction methods mentioned in Section \ref{spat:intro} all predicts the local condition with local historical climate data-based models. Data availability is an inevitable factor to consider when constructing these models. Some other domains using local historical data could also be impacted by data availability. Therefore, these domains could also benefit from the proposed method in this article. Domains such as forest fire forecasting \citep{fire2030043} and soil property prediction \citep{RYAN2000139} that are evolving with spatial methods could be further explored with a varied version of the proposed method. Other fields such as rain forecasting and atmospheric weather forecasting \citep{8819643} could implement the proposed method as an exploration of a fault-tolerance design.

\section{Conclusion}
\label{spat:conclude}
This article proposes a spatial interpolation-based frost prediction method. This method aims to eliminate the dependency of on-site historical datasets/sensors during model training, validation, testing, and future operations. The climate data is from weather stations in the NSW and ACT areas of Australia. Climate data from 75 weather stations are obtained. DEM and NDVI datasets of the study area are also utilized. The proposed methods are ensemble learning methods based on ANN models. The two ensemble methods are averaging and weighted averaging of weak predictors. Each weak predictor is constructed using one weather station as the climate data source. These models are also constructed with five-fold validation with the weather station data divided into five testing data folds. For each testing data fold, the other weather stations are used as training data sources. The baseline models are ANN models trained with on-site historical data. There are three experiments conducted to test the performance of models. The first experiment compares the raster map outputs of spatial interpolation-based models. Raster maps are constructed from the individual weak predictors, averaged results of weak predictors, and weighted averaged results of weak predictors. The results of the T-tests show that the raster maps for different folds are significantly different from each other. This shows that the models are significantly affected by the training datasets, hence, the division of folds is important. The second experiment shows that the weighted averaged method provides the lowest error among the spatial interpolation-based methods. After that, the final experiment reveals the effect of the number of available stations. Accuracy increases as the number of available stations increases. However, the event capture rate increases with the reduction of station number. This increase is related to the increase of false positives. Apparently, higher errors induce the triggering of more events, which increases both event capture rate and false positives. The experiments indicate three limitations of the work. The first limitation is accuracy. Accuracy needs to be increased to eliminate the relationship between high event capture rate and high false positives. Also, as a spatial interpolation-based method constructed by five-fold validation, only 15 stations per fold act as the ground truth to test the models. The models are also highly sensitive to the choice of training weather stations. This uncovers the importance of fold selection. Finally, the incorporation of time series prediction methods is defined as a future solution to extend the limited prediction window of the proposed method. In conclusion, the proposed spatial interpolation-based frost prediction method could capture frost events and be deployed as an alternative when on-site historical datasets are temporarily unavailable.

\section*{Data availability}
\subsection{Data Sources}
The weather station datasets can be obtained from the Australian Bureau of Meteorology (BOM) website \citep{bom2020datadir}. The weather stations with the following IDs are required: 70351, 70217, 65103, 66194, 68192, 75041, 66037, 63291, 73138, 51049, 62100, 58198, 67113, 61078, 61363, 69148, 61287, 51161, 74148, 58208, 66161, 62101, 47048, 65068, 69139, 59007, 58214, 60141, 68262, 66137, 58012, 75019, 56238, 63292, 49000, 67105, 63303, 58077, 68257, 66212, 55202, 68242, 74258, 65111, 58212, 70330, 48245, 54038, 72160, 72162, 72161, 50017, 60139, 61375, 68072, 68239, 61425, 46012, 64017, 69128, 68228, 67108, 69137, 52088, 61392, 67119, 55325, 61055, 50137, 69138, 61366, 65070, 61260, 69147, 68241. The raw DEM dataset is available on \citep{STRMDEM}. The NDVI datasets are available from \citep{MODISNDVI}. Boundary data of NSW and ACT are available from \citep{NSWBound} and \citep{ACTBound}.

The models are trained using the Tensorflow library (\url{https://www.tensorflow.org/}) in the Anaconda environment (\url{https://www.anaconda.com/}). The training code is available at \url{https://github.com/izho1042/Frost-Spatial-Prediction}.

\section*{Acknowledgment}
We would like to acknowledge the support of Food Agility CRC Ltd, 175 Pitt St., Sydney, NSW, 2000, Australia, funded under the Commonwealth Government CRC Program. The CRC Program supports industry-led collaborations between industry, researchers and the community.

The authors are with Radio Frequency and Communication Technologies (RFCT) research laboratory, Faculty of Engineering and IT, University of Technology Sydney, Ultimo, NSW 2007, Australia.

\bibliography{mybibfile}
\newpage
\appendix
\section{Raster Maps Generated from Models Created by Different Weather Stations}
\label{appx:spat_compare}

\begin{figure}[hbt!]%[!htpb]
\centering
\includegraphics[width=6.4in]{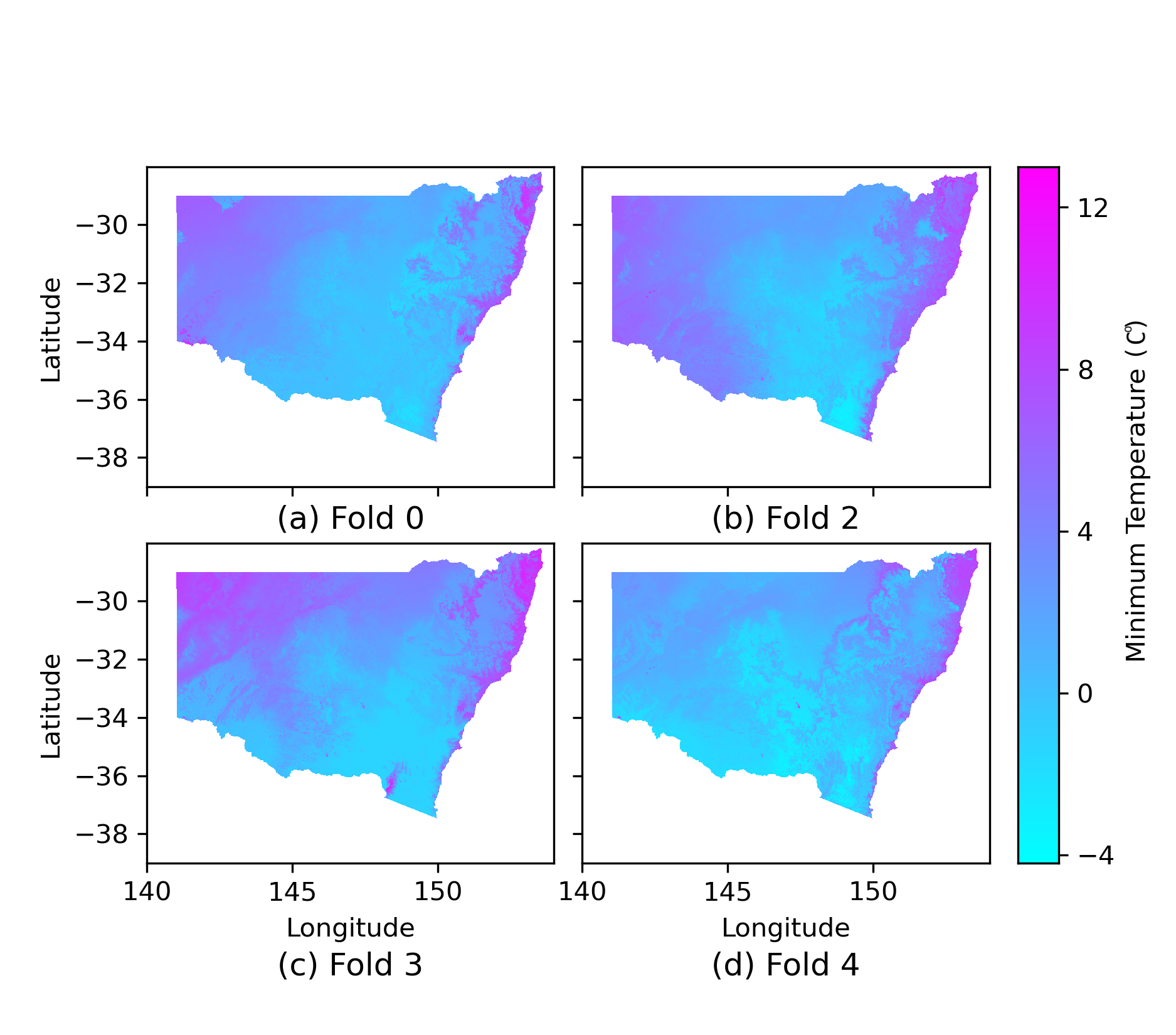} % 5.0
\DeclareGraphicsExtensions{.pdf,.jpeg,.png,}
\caption{Raster Maps from Models Using Weather Station 66137 as Climate Data Source Trained with the Datasets from Folds 0, 2, 3, and 4.}
\label{fig:spat_raster_66137}
\end{figure}

\begin{figure}[hbt!]%[!htpb]
\centering
\includegraphics[width=6.4in]{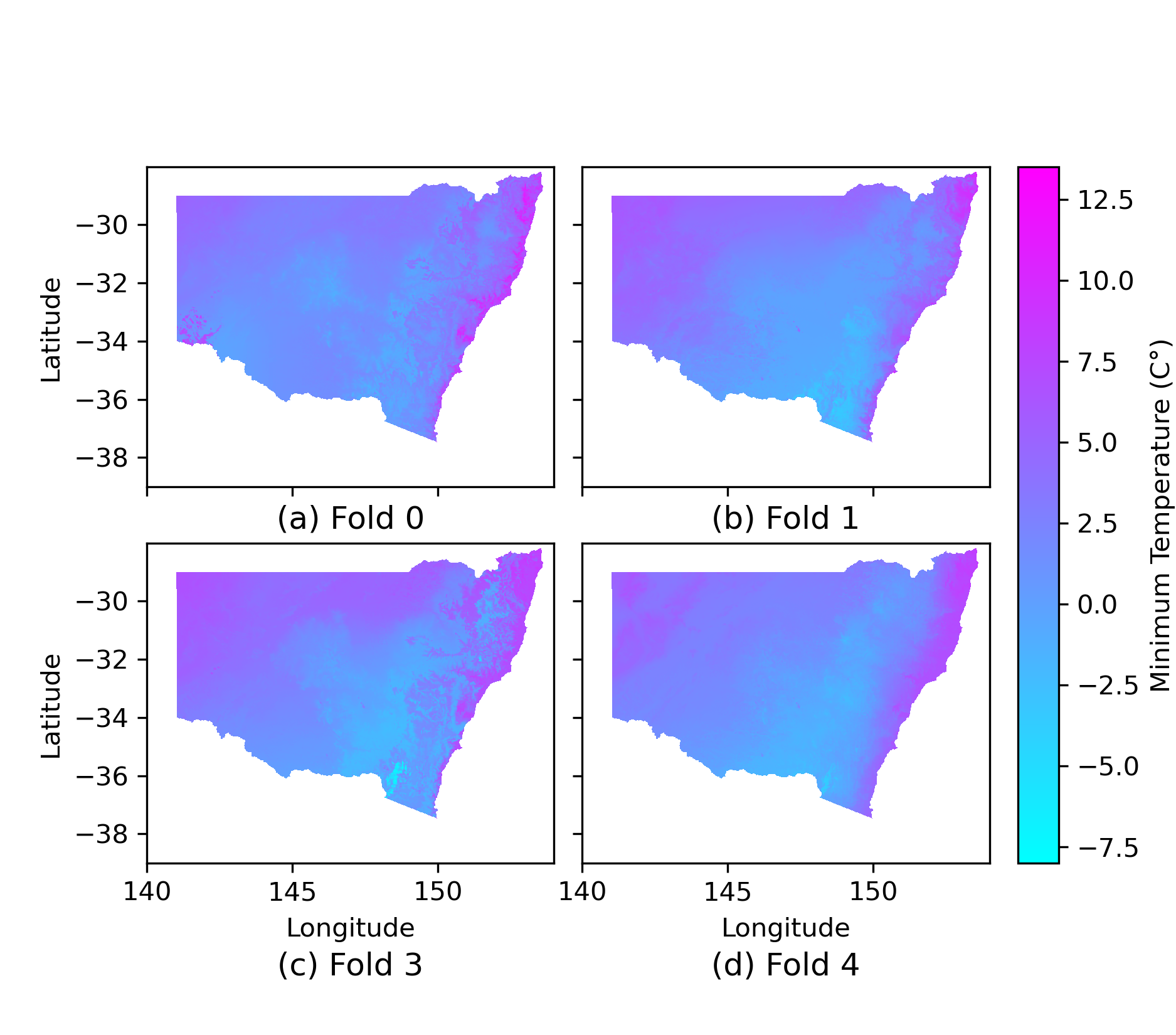}
\DeclareGraphicsExtensions{.pdf,.jpeg,.png,}
\caption{Raster Maps from Models Using Weather Station 58212 as Climate Data Source Trained with the Datasets from Folds 0, 1, 3, and 4.}
\label{fig:spat_raster_58212}
\end{figure}

\begin{figure}[hbt!]%[!htpb]
\centering
\includegraphics[width=6.4in]{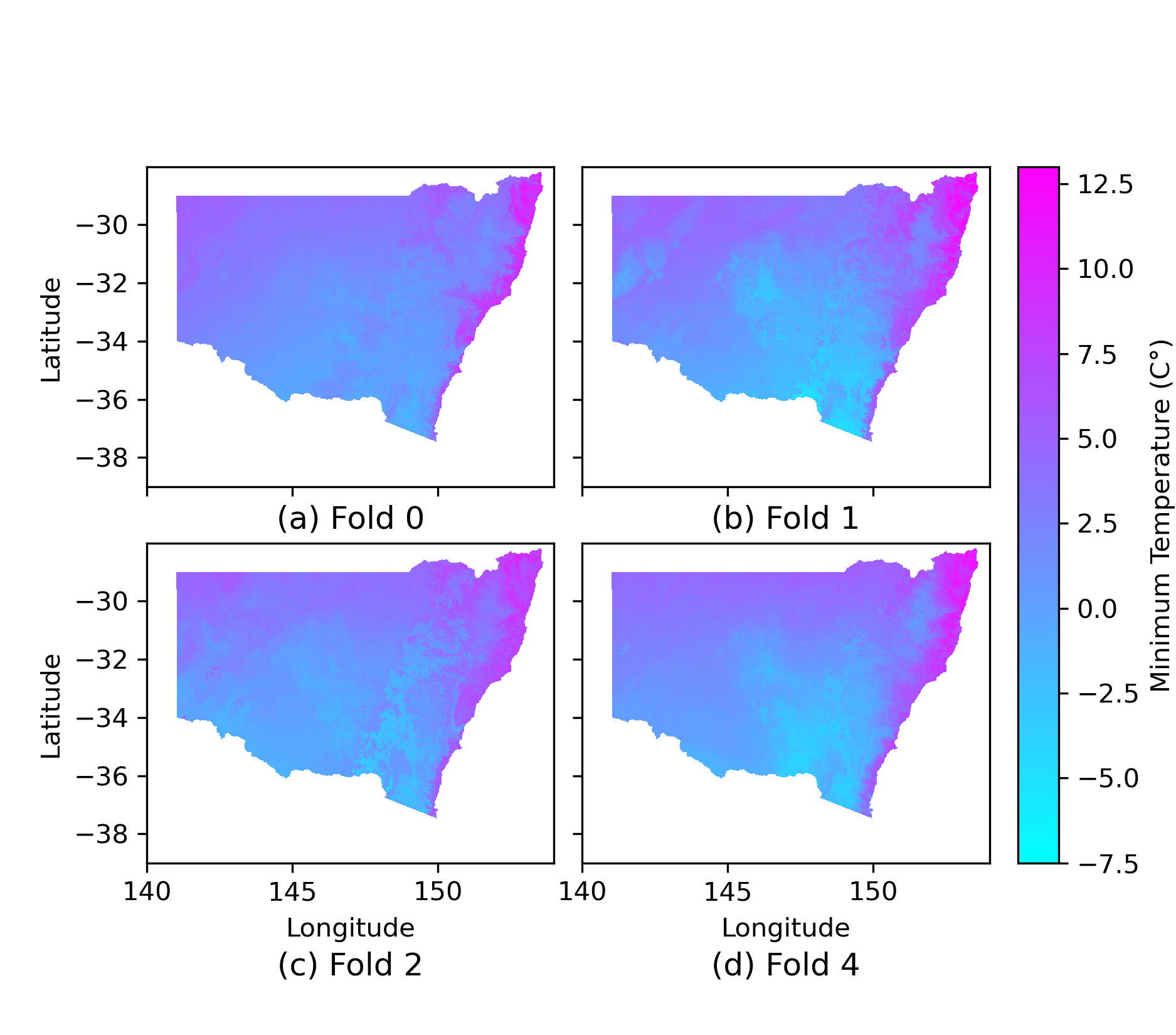}
\DeclareGraphicsExtensions{.pdf,.jpeg,.png,}
\caption{Raster Maps from Models Using Weather Station 72160 as Climate Data Source Trained with the Datasets from Folds 0, 1, 2, and 4.}
\label{fig:spat_raster_72160}
\end{figure}

\begin{figure}[hbt!]%[!htpb]
\centering
\includegraphics[width=6.4in]{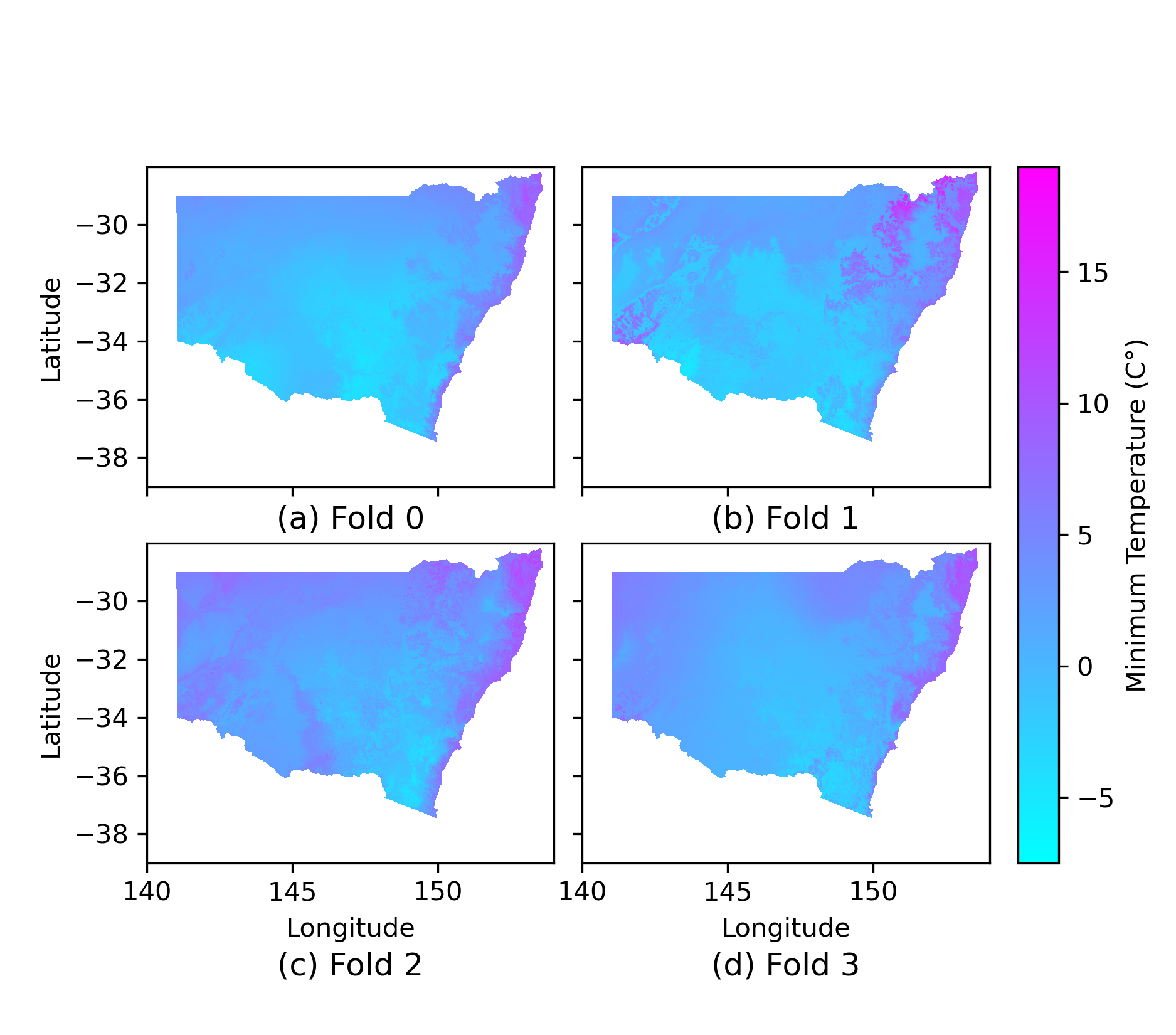}
\DeclareGraphicsExtensions{.pdf,.jpeg,.png,}
\caption{Raster Maps from Models Using Weather Station 67119 as Climate Data Source Trained with the Datasets from Folds 0--3.}
\label{fig:spat_raster_67119}
\end{figure}

\begin{figure}[hbt!]%[!htpb]
\centering
\includegraphics[width=6.4in]{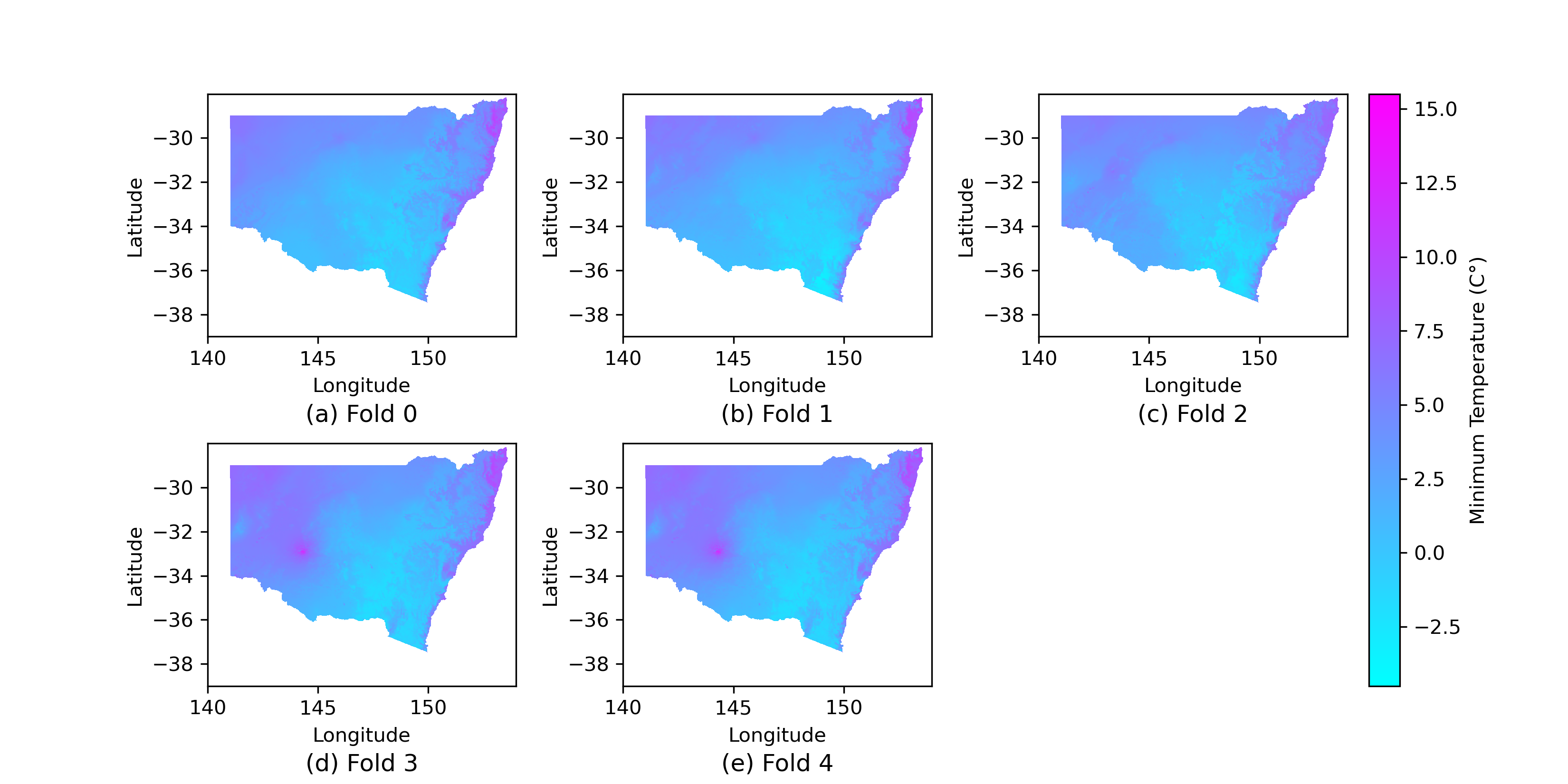}
\DeclareGraphicsExtensions{.pdf,.jpeg,.png,}
\caption{Raster Maps from Weighted Averaged Results per Fold.}
\label{fig:spat_raster_weighted}
\end{figure}
\FloatBarrier
\section{P-value Matrices Comparing Raster Map Results Generated from different folds}
\label{appx:spat_ttest}
\begin{table}[hbt!]%[!htbp]
%% increase table row spacing, adjust to taste
\renewcommand{\arraystretch}{1.3}
% if using array.sty, it might be a good idea to tweak the value of
%\extrarowheight as needed to properly center the text within the cells
\caption{P-value Matrix for Comparing Raster Map Results of Models Using Weather Station 66137 as Climate Data Source and Trained with the Datasets from Folds 0, 2, 3, 4}
\label{tab:spat_66137}
\centering
%% Some packages, such as MDW tools, offer better commands for making tables
%% than the plain LaTeX2e tabular which is used here.
\begin{tabular}{|p{\dimexpr 0.18\linewidth-2\tabcolsep}| p{\dimexpr 0.18\linewidth-2\tabcolsep} | p{\dimexpr 0.18\linewidth-2\tabcolsep}|p{\dimexpr 0.18\linewidth-2\tabcolsep}|p{\dimexpr 0.18\linewidth-2\tabcolsep}|}
\hline
& \textbf{Fold 0} & \textbf{Fold 2} & \textbf{Fold 3} & \textbf{Fold 4}\\
\hline
\textbf{Fold 0}&N/A& 0.00& 0.00&0.00\\\hline
\textbf{Fold 2}&0.00& N/A& 0.00&0.00\\\hline
\textbf{Fold 3}&0.00& 0.00& N/A&0.00\\\hline
\textbf{Fold 4}&0.00& 0.00& 0.00&N/A
\\\hline
\end{tabular}
\end{table}

\begin{table}[hbt!]%[!htbp]
%% increase table row spacing, adjust to taste
\renewcommand{\arraystretch}{1.3}
% if using array.sty, it might be a good idea to tweak the value of
%\extrarowheight as needed to properly center the text within the cells
\caption{P-value Matrix for Comparing Raster Map Results of Models Using Weather Station 58212 as Climate Data Source and Trained with the Datasets from Folds 0, 1, 3, 4}
\label{tab:spat_58212}
\centering
%% Some packages, such as MDW tools, offer better commands for making tables
%% than the plain LaTeX2e tabular which is used here.
\begin{tabular}{|p{\dimexpr 0.18\linewidth-2\tabcolsep}| p{\dimexpr 0.18\linewidth-2\tabcolsep} | p{\dimexpr 0.18\linewidth-2\tabcolsep}|p{\dimexpr 0.18\linewidth-2\tabcolsep}|p{\dimexpr 0.18\linewidth-2\tabcolsep}|}
\hline
& \textbf{Fold 0} & \textbf{Fold 1} & \textbf{Fold 3} & \textbf{Fold 4}\\
\hline
\textbf{Fold 0}&N/A& 8.83e-206& 0.00&0.00\\\hline
\textbf{Fold 1}&8.83e-206& N/A& 0.00&0.00\\\hline
\textbf{Fold 3}&0.00& 0.00& N/A&0.00\\\hline
\textbf{Fold 4}&0.00& 0.00& 0.00&N/A
\\\hline
\end{tabular}
\end{table}

\begin{table}[hbt!]%[!htbp]
%% increase table row spacing, adjust to taste
\renewcommand{\arraystretch}{1.3}
% if using array.sty, it might be a good idea to tweak the value of
%\extrarowheight as needed to properly center the text within the cells
\caption{P-value Matrix for Comparing Raster Map Results of Models Using Weather Station 72160 as Climate Data Source and Trained with the Datasets from Folds 0, 1, 2, 4}
\label{tab:spat_72160}
\centering
%% Some packages, such as MDW tools, offer better commands for making tables
%% than the plain LaTeX2e tabular which is used here.
\begin{tabular}{|p{\dimexpr 0.18\linewidth-2\tabcolsep}| p{\dimexpr 0.18\linewidth-2\tabcolsep} | p{\dimexpr 0.18\linewidth-2\tabcolsep}|p{\dimexpr 0.18\linewidth-2\tabcolsep}|p{\dimexpr 0.18\linewidth-2\tabcolsep}|}
\hline
& \textbf{Fold 0} & \textbf{Fold 1} & \textbf{Fold 2} & \textbf{Fold 4}\\
\hline
\textbf{Fold 0}&N/A& 0.00& 0.00&0.00\\\hline
\textbf{Fold 1}&0.00& N/A& 0.00&6.91e-22\\\hline
\textbf{Fold 2}&0.00& 0.00& N/A&0.00\\\hline
\textbf{Fold 4}&0.00& 6.91e-22& 0.00&N/A
\\\hline
\end{tabular}
\end{table}

\begin{table}[hbt!]%[!htbp]
%% increase table row spacing, adjust to taste
\renewcommand{\arraystretch}{1.3}
% if using array.sty, it might be a good idea to tweak the value of
%\extrarowheight as needed to properly center the text within the cells
\caption{P-value Matrix for Comparing Raster Map Results of Models Using Weather Station 67119 as Climate Data Source and Trained with the Datasets from Folds 0--3}
\label{tab:spat_67119}
\centering
%% Some packages, such as MDW tools, offer better commands for making tables
%% than the plain LaTeX2e tabular which is used here.
\begin{tabular}{|p{\dimexpr 0.18\linewidth-2\tabcolsep}| p{\dimexpr 0.18\linewidth-2\tabcolsep} | p{\dimexpr 0.18\linewidth-2\tabcolsep}|p{\dimexpr 0.18\linewidth-2\tabcolsep}|p{\dimexpr 0.18\linewidth-2\tabcolsep}|}
\hline
& \textbf{Fold 0} & \textbf{Fold 1} & \textbf{Fold 2} & \textbf{Fold 3}\\
\hline
\textbf{Fold 0}&N/A& 0.00& 0.00&0.00\\\hline
\textbf{Fold 1}&0.00& N/A& 0.00&0.00\\\hline
\textbf{Fold 2}&0.00& 0.00& N/A&0.00\\\hline
\textbf{Fold 3}&0.00& 0.00& 0.00&N/A
\\\hline
\end{tabular}
\end{table}

\begin{table}[hbt!]%[!htbp]
%% increase table row spacing, adjust to taste
\renewcommand{\arraystretch}{1.3}
% if using array.sty, it might be a good idea to tweak the value of
%\extrarowheight as needed to properly center the text within the cells
\caption{P-value Matrix for Comparing Weighted Averaged Raster Map Results of Each Fold}
\label{tab:spat_weighted}
\centering
%% Some packages, such as MDW tools, offer better commands for making tables
%% than the plain LaTeX2e tabular which is used here.
\begin{tabular}{|p{\dimexpr 0.14\linewidth-2\tabcolsep}| p{\dimexpr 0.14\linewidth-2\tabcolsep} | p{\dimexpr 0.14\linewidth-2\tabcolsep}|p{\dimexpr 0.18\linewidth-2\tabcolsep}|p{\dimexpr 0.14\linewidth-2\tabcolsep}|p{\dimexpr 0.18\linewidth-2\tabcolsep}|}
\hline
& \textbf{Fold 0} & \textbf{Fold 1} & \textbf{Fold 2} & \textbf{Fold 3}& \textbf{Fold 4}\\
\hline
\textbf{Fold 0}&N/A&0.00&0.00&0.00&0.00\\\hline
\textbf{Fold 1}&0.00& N/A& 0.00&0.00&0.00\\\hline
\textbf{Fold 2}&0.00& 0.00& N/A&0.00&0.00\\\hline
\textbf{Fold 3}&0.00& 0.00& 0.00&N/A&0.00\\\hline
\textbf{Fold 4}&0.00& 0.00& 0.00&0.00&N/A
\\\hline
\end{tabular}
\end{table}

\end{document}